%% file: main.tex
\documentclass[a4paper,11pt]{article}
\usepackage[colorlinks = true,
            linkcolor = magenta,
            urlcolor  = blue,
            citecolor = green,
            anchorcolor = blue]{hyperref}
            
\usepackage{amsmath,amsfonts,amssymb,amsthm,bm,dsfont}
\usepackage{authblk}  
\usepackage[british]{babel}
\usepackage{bbm}  
\usepackage[sorting=none]{biblatex}
\usepackage{caption}
\usepackage{color}
\usepackage{enumitem}
\usepackage[T1]{fontenc}
\usepackage[top=2.5cm, bottom=2.5cm, left=2.5cm, right=2.5cm]{geometry}
\usepackage[pdftex]{graphicx}
\usepackage{graphbox}       
\PassOptionsToPackage{hyphens}{url}
\usepackage{hyperref}
\usepackage[utf8]{inputenc}
\usepackage{csquotes}  
\usepackage[mono=false]{libertine}
\usepackage{nicefrac}
\usepackage{mathtools}
\usepackage{microtype}      
\usepackage{subcaption}
\usepackage{xcolor}
\usepackage{booktabs} 
\usepackage{siunitx}
\usepackage{array, booktabs, makecell}
\usepackage{multirow}

\input{math_commands}

\newcommand{\citet}[1]{\cite{#1}}
\newcommand{\citep}[1]{\cite{#1}}
\definecolor{C0}{HTML}{1f77b4}
\definecolor{C1}{HTML}{ff7f0e}
\definecolor{C2}{HTML}{2ca02c}
\definecolor{C3}{HTML}{d62728}
\definecolor{C4}{HTML}{9467bd}
\definecolor{C5}{HTML}{8c564b}
\definecolor{C6}{HTML}{e377c2}
\definecolor{C7}{HTML}{7f7f7f}
\definecolor{C8}{HTML}{bcbd22}
\definecolor{C9}{HTML}{17becf}

\bibliography{refs}

\title{\huge\textbf{End-to-end symbolic regression with transformers}}

\author[1,2]{Pierre-Alexandre Kamienny\thanks{pakamienny@fb.com}$^\dagger$}
\author[1,3]{St\'ephane d'Ascoli\thanks{stephane.dascoli@gmail.com}$^\dagger$}
\author[1]{Guillaume Lample}
\author[1]{François Charton}
\affil[1]{Meta AI, Paris}
\affil[2]{ISIR MLIA, Sorbonne Université, Paris}
\affil[3]{Department of Physics, Ecole Normale Sup\'erieure, Paris}
\date{}

\begin{document}

\maketitle
\def\thefootnote{$\dagger$}\footnotetext{Equal contribution.}\def\thefootnote{\arabic{footnote}}

\input{abstract}

\input{main_text}

\clearpage
\printbibliography

\clearpage
\appendix
\input{appendix}

\end{document}

%% file: math_commands.tex
\DeclareUnicodeCharacter{2212}{-}

\usepackage{xcolor}
\definecolor{C0}{HTML}{1f77b4}
\definecolor{C1}{HTML}{ff7f0e}
\definecolor{C2}{HTML}{2ca02c}
\definecolor{C3}{HTML}{d62728}
\definecolor{C4}{HTML}{9467bd}
\definecolor{C5}{HTML}{8c564b}
\definecolor{C6}{HTML}{e377c2}
\definecolor{C7}{HTML}{7f7f7f}
\definecolor{C8}{HTML}{bcbd22}
\definecolor{C9}{HTML}{17becf}

\usepackage[normalem]{ulem}

\newcommand{\din}{D}

\newcommand{\demb}{d_\text{emb}}

\newcommand{\ntest}{N_\text{test}}
\newcommand{\numbags}{B}
\newcommand{\numrefs}{K}
\newcommand{\numbeam}{C}

%% file: abstract.tex
\begin{abstract}
Symbolic regression, the task of predicting the mathematical expression of a function from the observation of its values, is a difficult task which usually involves a two-step procedure: predicting the "skeleton" of the expression up to the choice of numerical constants, then fitting the constants by optimizing a non-convex loss function. The dominant approach is genetic programming, which evolves candidates by iterating this subroutine a large number of times. Neural networks have recently been tasked to predict the correct skeleton in a single try, but remain much less powerful.

In this paper, we challenge this two-step procedure, and task a Transformer to directly predict the full mathematical expression, constants included. One can subsequently refine the predicted constants by feeding them to the non-convex optimizer as an informed initialization. We present ablations to show that this end-to-end approach yields better results, sometimes even without the refinement step. We evaluate our model on problems from the SRBench benchmark and show that our model approaches the performance of state-of-the-art genetic programming with several orders of magnitude faster inference. 

\end{abstract}

%% file: main_text.tex
\section*{Introduction}

Inferring mathematical laws from experimental data is a central problem in natural science; having observed a variable $y$ at $n$ points $\{x_i\}_{i\in \mathbb{N}_n}$, it implies finding a function $f$ such that $y_i \approx f(x_i)$ for all $i \in \mathbb{N}_n$. Two types of approaches exist to solve this problem. In \emph{parametric statistics} (PS), the function $f$ is defined by a small number of parameters that can directly be estimated from the data. On the other hand, \emph{machine learning} (ML) techniques such as decision trees and neural networks select $f$ from large families of non-linear functions by minimizing a loss over the data. The latter relax the assumptions about the underlying law, but their solutions are more difficult to interpret, and tend to overfit small experimental data sets, yielding poor extrapolation performance.

Symbolic regression (SR) stands as a middle ground between PS and ML approaches: $f$ is selected from a large family of functions, but is required to be defined by an interpretable analytical expression.
It has already proved extremely useful in a variety of tasks such as inferring physical laws~\cite{udrescu2020ai,Cranmer2020DiscoveringSM}.

SR is usually performed in two steps. First, predicting a ``skeleton'', a parametric function using a pre-defined list of operators -- typically, the basic operations ($+,\times,\div$) and functions ($\operatorname{sqrt},\exp, \sin$, etc.). It determines the general shape of the law up to a choice of constants, e.g. $f(x) = \cos(ax+b)$. Then, the constants in the skeleton ($a,b$) are estimated using optimization techniques, typically the Broyden–Fletcher–Goldfarb–Shanno algorithm (BFGS). 

The leading algorithms for SR rely on genetic programming (GP). At each generation, a population of candidates is predicted, and the fittest ones are selected based on the data, and mutated to build the next generation. The algorithm iterates this procedure until a satisfactory level of accuracy is achieved.

\begin{figure}
    \includegraphics[width=\linewidth]{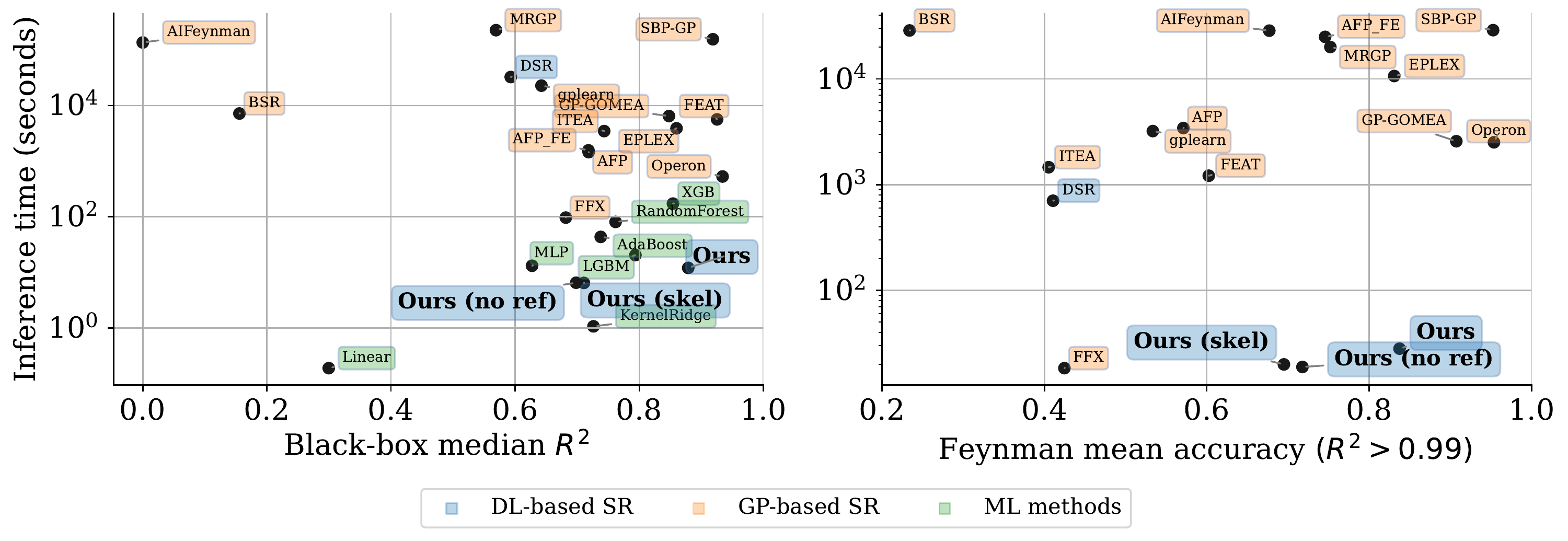}
  \caption{\textbf{Our model outperforms previous DL-based methods and offers at least an order of magnitude inference speedup compared to SOTA GP-based methods.} Pareto plot comparing the average test performance and inference time of our models with baselines provided by the SRbench benchmark~\cite{la2021contemporary}, both on Feynman SR problems~\cite{udrescu2020ai} and black-box regression problems. We use colors to distinguish three families of models: \textbf{\textcolor{C0}{deep-learning based SR}},  \textbf{\textcolor{C1}{genetic programming-based SR}} and  \textbf{\textcolor{C2}{classic machine learning methods}} (which do not provide an interpretable solutions). A similar Pareto plot against formula complexity is provided in Fig.~\ref{fig:pareto-size}.}
  \label{fig:pareto}
  \end{figure}

While GP algorithms achieve good prediction accuracy, they are notably slow (see the Pareto plot of Fig.~\ref{fig:pareto}). 
Indeed, the manually predefined function space to search is generally vast, and each generation involves a costly call to the BFGS routine. Also, GP does not leverage past experience: every new problem is learned from scratch. This makes GP techniques inapplicable to situations where fast computation is needed, for instance in reinforcement learning and physics environments~\cite{garnelo2016towards,landajuela2021discovering}.  

Pre-training neural networks built for language modelling on large datasets of synthetic examples has recently been proposed for SR~\cite{valipour2021symbolicgpt,biggio2021neural}. These references follow the two-step procedure (predicting the skeleton then fitting the constants) inherited from GP.
Once the model is pre-trained, the skeleton is predicted via a simple forward pass, and a single call to BFGS is needed, thus resulting in a significant speed-up compared to GP.
However, these methods are not as accurate as state-of-the-art GP, and have so far been limited to low-dimensional functions ($\din\leq 3$). We argue that two reasons underlie their shortcomings. 

First, skeleton prediction is an ill-posed problem that does not provide sufficient supervision: different instances of the same skeleton can have very different shapes, and instances of very different skeletons can be very close. 
Second, the loss function minimized by BFGS can be highly non-nonconvex: even when the skeleton is perfectly predicted, the correct constants are not guaranteed to be found.
For these reasons, we believe, and will show, that doing away with skeleton estimation as a intermediary step can greatly facilitate the task of SR for language models.

\paragraph{Contributions}

In this paper, we train Transformers over synthetic datasets to perform \textbf{end-to-end (E2E)} symbolic regression: solutions are predicted directly, without resorting to skeletons. To this effect, we leverage a hybrid \textbf{symbolic-numeric vocabulary}, that uses both symbolic tokens for the operators and variables and numeric tokens for the constants. 
One can then perform a \textbf{refinement} of the predicted constants by feeding them as informed guess to BFGS, mitigating non-linear optimization issues. Finally, we introduce \textbf{generation and inference techniques} that allow our models to scale to larger problems: up to $10$ input features against $3$ in concurrent works.

Evaluated over the SRBench benchmark \cite{la2021contemporary}, our model significantly narrows the accuracy gap with state-of-the-art GP techniques, while providing several orders of magnitude of inference time speedup (see Fig.~\ref{fig:pareto}). We also demonstrate strong robustness to noise and extrapolation capabilities. 

Finally, we will provide an online demonstration of our model at \url{https://bit.ly/3niE5FS} and will open-source our implementation as a Scikit-learn compatible regressor at the following address: \url{https://github.com/facebookresearch/symbolicregression}.

\paragraph{Related work}

SR is a challenging task that traces back from a few decades ago, with a large number of open-source and commercial softwares, and has already been used to accelerate scientific discoveries~\cite{Archiga2021AcceleratingUO,Udrescu2021SymbolicPD,Butter2021BackTT}. Most popular frameworks for symbolic regression use GP~\cite{schmidt2011age,schmidt2009distilling,la2018learning,mcconaghy2011ffx,virgolin2021improving,de2021interaction,arnaldo2014multiple,virgolin2019linear,kommenda2020genetic} (see~\cite{la2021contemporary} for a recent review), but SR has also seen growing interest from the Deep Learning (DL) community, motivated by the fact that neural networks are good at identifying qualitative patterns.  

Neural networks have  been combined with GP algorithms, e.g. to simplify the original dataset~\cite{udrescu2020ai}, or to propose a good starting distribution over mathematical expressions\cite{petersen2019deep}. \cite{martius2016extrapolation,sahoo2018learning} propose modifications to feed-forward networks to include interpretable components, i.e. replacing usual activation functions by operators such as $\cos,\sin$, however these are hard to optimize and prone to numerical issues. 

Language models, and especially Transformers~\cite{vaswani2017attention}, have been trained over synthetic datasets to solve various mathematical problems: integration~\cite{lample2019deep}, dynamical systems~\cite{charton2020learning}, linear algebra~\cite{charton2021linear}, formal logic~\cite{hahn2020teaching} and theorem proving~\cite{polu2020generative}. A few papers apply these techniques to symbolic regression: the aforementioned references~\cite{biggio2021neural,valipour2021symbolicgpt} train Transformers to predict function skeletons, while \cite{d2022deep} infers one-dimensional recurrence relations in sequences of numbers. 

The recently introduced SRBench~\cite{la2021contemporary} provides a benchmark for rigorous evaluation of SR methods, in addition to $14$ SR methods and $7$ ML baselines which we will compare to in this work.

\section{Data generation} \label{sec:generation}

Our approach consists in pre-training language models on vast synthetic datasets. Each training example is a pair: a set of $N$ points $(x,y)\in\mathbb{R}^D\times \mathbb{R}$ as the input, and a function $f$ such that $y=f(x)$ as the target\footnote{We only consider functions from $\mathbb{R}^D$ into $\mathbb{R}$; the general case $f:\mathbb{R}^D \to \mathbb{R}^{P}$ can be handled as $P$ independent subproblems.} Examples are generated by first sampling a random function $f$, then a set of $N$ input values $(x_i)_{i\in \mathbb{N}_N}$ in $\mathbb{R}^D$, and computing $y_i=f(x_i)$. 

\subsection{Generating functions}
\label{sec:func-generation}

To sample functions $f$, we follow the seminal approach of Lample and Charton~\cite{lample2019deep}, and generate random trees with mathematical operators as internal nodes and variables or constants as leaves. The procedure is detailed below (see Table~\ref{tab:generator} in the Appendix for the values of parameters):
\begin{enumerate}[leftmargin=1cm,noitemsep]
    \item Sample the desired \textbf{input dimension} $\din$ of the function $f$ from $\mathcal{U}\{1, D_\text{max}\}$.
    \item Sample the number of \textbf{binary operators} $b$ from $\mathcal{U}\{ D-1,  D+b_\text{max}\}$ then sample $b$ operators from $\mathcal U\{+,-,\times\}$\footnote{Note that although the division operation is technically a binary operator, it appears much less frequently than additions and multiplications in typical expressions~\cite{guimera2020bayesian}, hence we replace it by the unary operator $\operatorname{inv:} x\to1/x$.}. 
    \item Build a \textbf{binary tree} with those $b$ nodes, using the sampling procedure of~\cite{lample2019deep}. 
    \item For each \textbf{leaf} in the tree, sample one of the variables $x_d$, $d \in \mathbb{N}_\din$.
    \item Sample the number of \textbf{unary operators} $u$ from $\mathcal{U}\{0, u_\text{max}\}$ then sample $u$ operators from the list $O_u$ in Table~\ref{tab:generator}, and insert them at random positions in the tree. 
    \item For each variable $x_d$ and unary operator $u$, apply a random \textbf{affine transformation}, i.e. replace $x_d$ by $ax_d + b$, and $u$ by $au+b$, with $(a,b)$ sampled from $\mathcal D_\text{aff}$. 
\end{enumerate}

Note that since we require independent control on the number of unary operators (which is independent of $D$) and binary operators (which depends on $D$), we cannot directly sample a unary-binary tree as in~\cite{lample2019deep}. Note also that the first $D$ variables are sampled in ascending order to obtain the desired input dimension, which means functions with missing variables such as $x_1+x_3$ are never encountered; this is not an issue as our model can always set the prefactor of $x_2$ to zero. 
As discussed quantitatively in App.~\ref{app:memorization}, the number of possible skeletons as well as the random sampling of numerical constants guarantees that our model almost never sees the same function twice, and cannot simply perform memorization.

\begin{figure*}[tb]
    \centering
    \includegraphics[width=\linewidth]{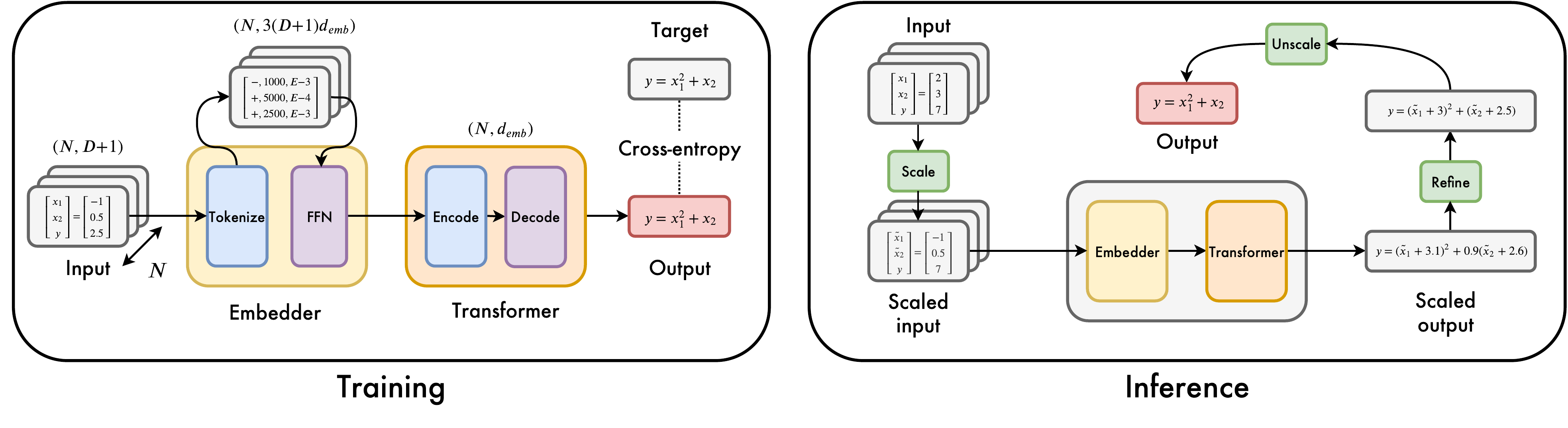}
    \caption{\textbf{Sketch of our model.} During training, the inputs are all whitened. At inference, we whiten them as a pre-processing step; the predicted function must then be unscaled to account for the whitening.}
    \label{fig:sketch}
\end{figure*}

\subsection{Generating inputs}
\label{sec:inputs}

For each function $f:\mathbb{R}^\din\to\mathbb{R}$, we sample $N \in \mathcal{U}\{10\din, N_\text{max}\}$ \textbf{input values} $x_i \in \mathbb{R}^\din$ from the distribution $\mathcal D_{x}$ described below, and compute the corresponding \textbf{output values} $y_i=f(x_i)$. 
If any $x_i$ is outside the domain of definition of $f$ or if any $y_i$ is larger $10^{100}$, the process is aborted, and we start again by generating a new function. Note that rejecting and resampling out-of-domain values of $x_i$, the obvious and cheaper alternative, would provide the model with additional information about $f$, by allowing it to learn its domain of definition.

To maximize the diversity of input distributions seen at training time, we sample our inputs from a mixture of distributions (uniform or gaussian), centered around $k$ random centroids\footnote{For $k\to\infty$, such a mixture could in principe approximate any input distribution.}, see App.~\ref{app:data} for some illustrations at $\din=2$. Input samples are generated as follows:
\begin{enumerate}[leftmargin=1cm, noitemsep]
    \item Sample a \textbf{number of clusters} $k \sim \mathcal{U}\{1, k_{max}\}$ and $k$ \textbf{weights} $w_i \sim \mathcal{U}(0,1)$, which are then normalized so that $\sum_i w_i=1$.
    \item For each cluster $i\in \mathbb{N}_k$, sample a \textbf{centroid} $\mu_i \sim \mathcal{N}(0,1)^\din$, a vector of \textbf{variances} $\sigma_i\sim \mathcal{U}(0,1)^D$ and a \textbf{distribution shape} (gaussian or uniform) $\mathcal{D}_i \in \{\mathcal{N}, \mathcal{U}\}$.
    \item For each cluster $i\in \mathbb{N}_k$, sample $\lfloor w_i N\rfloor$ \textbf{input points} from $\mathcal{D}_i(\mu_i, \sigma_i)$ then apply a \textbf{random rotation} sampled from the Haar distribution.
    \item Finally, \textbf{concatenate} all the points obtained and \textbf{whiten} them by substracting the mean and dividing by the standard deviation along each dimension.
\end{enumerate}

\subsection{Tokenization}

Following~\citet{charton2021linear}, we represent numbers in base $10$ floating-point notation, round them to four significant digits, and encode them as sequences of $3$ tokens: their sign, mantissa (between \texttt{0} and \texttt{9999}), and exponent (from \texttt{E-100} to \texttt{E100}). 

To represent mathematical functions as sequences, we enumerate the trees in prefix order, i.e. direct Polish notation, as in~\cite{lample2019deep}: operators and variables and integers are represented as single autonomous tokens, and constants are encoded as explained above.

For example, the expression $f(x)=\cos(2.4242 x)$ is encoded as $\texttt{[cos,mul,+,2424,E-3,x]}$. Note that the vocabulary of the decoder contains a mix of symbolic tokens (operators and variables) and numeric tokens, whereas that of the encoder contains only numeric tokens\footnote{The embeddings of numeric tokens are \emph{not} shared between the encoder and decoder.}.

\section{Methods}

Below we describe our approach for end-to-end symbolic regression; please refer to Fig.~\ref{fig:sketch} for an illustration.

\subsection{Model}

\paragraph{Embedder}
Our model is provided $N$ input points $(x,y) \in \mathbb{R}^{\din+1}$, each of which is represented as $3(\din+1)$ tokens of dimension $\demb$. As $D$ and $N$ become large, this results in long input sequences (e.g. $6600$ tokens for $\din=10$ and $N=200$), which challenge the quadratic complexity of Transformers. To mitigate this, we introduce an embedder to map each input point to a single embedding. 

The embedder pads the empty input dimensions to $D_\text{max}$, then feeds the 
$3(D_\text{max}+1)\demb$-dimensional vector into a 2-layer fully-connected feedforward network (FFN) with ReLU activations, which projects down to dimension $\demb$\footnote{We explored various architectures for the embedder, but did not obtain any improvement; this does not appear to be a critical part of the model.} The resulting $N$ embeddings of dimension $\demb$ are then fed to the Transformer.

\paragraph{Transformer}

We use a sequence to sequence Transformer architecture~\cite{vaswani2017attention} with 16 attention heads and an embedding dimension of 512, containing a total of 86M parameters. Like~\cite{charton2021linear}, we observe that the best architecture for this problem is asymmetric, with a deeper decoder: we use 4 layers in the encoder and 16 in the decoder. A notable property of this task is the permutation invariance of the $N$ input points. To account for this invariance, we remove positional embeddings from the encoder.

\begin{figure}[tb]
    \centering
    \includegraphics[width=\linewidth]{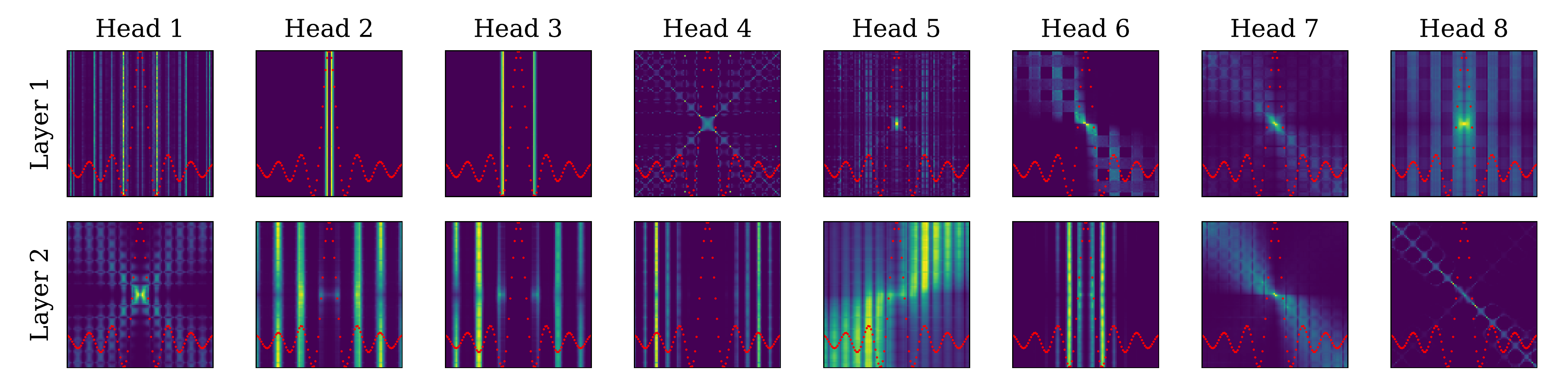}
    \caption{\textbf{Attention heads reveal intricate mathematical analysis.} We considered the expression $f(x)=\sin(x)/x$, with $N=100$ input points sampled between $-20$ and $20$ (red dots; the y-axis is arbitrary).  We plotted the attention maps of a few heads of the encoder, which are $N\times N$ matrices where the element $(i,j)$ represents the attention between point $i$ and point $j$. Notice that heads 2, 3 and 4 of the second layer analyze the periodicity of the function in a Fourier-like manner.}
    \label{fig:attention-small}
\end{figure}

As shown in Fig.~\ref{fig:attention-small} and detailed in App.~\ref{app:attention}, the encoder captures the most distinctive features of the functions considered, such as critical points and periodicity, and blends a mix of short-ranged heads focusing on local details with long-ranged heads which capture the global shape of the function.

\paragraph{Training}

We optimize a cross-entropy loss with the Adam optimizer, warming up the learning rate from $10^{-7}$ to $2.10^{-4}$ over the first 10,000 steps, then decaying it as the inverse square root of the number of steps, following \citep{vaswani2017attention}. We hold out a validation set of $10^4$ examples from the same generator, and train our models until the accuracy on the validation set saturates (around 50 epochs of 3M examples).

Input sequence lengths vary significantly with the number of points $N$; to avoid wasteful padding, we batch together examples of similar lengths, ensuring that a full batch contains a minimum of 10,000 tokens.
On 32 GPU with 32GB memory each, one epoch is processed in about half an hour.

\subsection{Inference tricks}

In this section, we describe three tricks to improve the performance of our model at inference. 

\paragraph{Refinement}

\begin{table}[htb]
    \centering
    \begin{tabular}{c|c}
    \toprule
        Model & Function $f(x,y)$ \\
        \midrule
        Target  & $\sin(10x) \exp(0.1y)$ \\
        Skeleton + BFGS            & $−\sin(1.7x)(0.059y+0.19)$ \\
        E2E no BFGS        & $\sin(9.9x) \exp(0.1y)$ \\
        E2E + BFGS random init    & $−\sin(0.095x) \exp(0.27y)$ \\
        E2E + BFGS model init           & $\sin(10x) \exp(0.1y)$ \\
    \bottomrule
    \end{tabular}
    \vspace{.4cm}
    \caption{\textbf{The importance of an end-to-end model with refinement.} The skeleton approach recovers an incorrect skeleton. The E2E approach predicts the right skeleton. Refinement worsens original prediction when randomly initialized, and yields the correct result when initialized with predicted constants.}
    \label{tab:refinement}
\end{table}

Previous language models for SR, such as \cite{biggio2021neural}, follow a \emph{skeleton} approach: they first predict equation skeletons, then fit the constants with a non-linear optimisation solver such as BFGS. In this paper, we follow an \emph{end-to-end} (E2E) approach: predicting simultaneously the function and the values of the constants. However, we improve our results by adding a \emph{refinement} step: fine-tuning the constants a posteriori with BFGS, initialized with our model predictions\footnote{To avoid BFGS having to approximate gradients via finite differences, we provide the analytical expression of the gradient using \textit{sympytorch} \cite{sympytorch} and \textit{functorch} \cite{functorch2021}.}.

This results in a large improvement over the skeleton approach, as we show by training a Transformer to predict skeletons in the same experimental setting. The improvement comes from two reasons: first, prediction of the full formula provides better supervision, and helps the model predict the skeleton; second, the BFGS routine strongly benefits from the informed initial guess, which helps the model predict the constants. This is illustrated qualitatively in Table~\ref{tab:refinement}, and quantitatively in Table~\ref{tab:in-domain}.

\paragraph{Scaling}
\label{sec:scaler}
As described in Section~\ref{sec:inputs}, all input points presented to the model during training are whitened: their distribution is centered around the origin and has unit variance. To allow accurate prediction for input points with a different mean and variance, we introduce a scaling procedure at inference time. Let $f$ the function to be inferred, $x$ be the input points, and $ \mu=\operatorname{mean}(x),  \sigma=\operatorname{std}(x)$. As illustrated in Fig.~\ref{fig:sketch} we pre-process the input data by replacing $x$ by 
$\tilde x = \frac{x- \mu}{ \sigma}$. The model then predicts $\hat f(\tilde x) = \hat f( \sigma x + \mu)$, and we can recover an approximation of $f$ by unscaling the variables in $\hat f$. 

This gives our model the desirable property to be insensitive to the scale of the input points: DL-based approaches to SR are known to fail when the inputs are outside the range of values seen during training~\cite{d2022deep,charton2021linear}. Note that here, the scale of the inputs translates to the scale of the constants in the function $f$; although these coefficients are sampled in $\mathcal{D}_\text{aff}$ during training, coefficients outside $\mathcal{D}_\text{aff}$ can be expressed by multiplication of constants in $\mathcal{D}_\text{aff}$. 

\paragraph{Bagging and decoding} 
Since our model was trained on $N\leq 200$ input points, it does not perform satisfactorily at inference when presented with more than $200$ input points. To take advantage of large datasets while accommodating memory constraints, we perform \emph{bagging}: whenever $N$ is larger than $200$ at inference, we randomly split the dataset into $\numbags$ bags of $200$ input points\footnote{Smarter splits, e.g. diversity-preserving, could be envisioned, but were not considered here.}. 

For each bag, we apply a forward pass and generate $\numbeam$ function candidates via random sampling or beam search using the next token distribution. As shown in App.~\ref{app:beam} (Fig.~\ref{fig:decoding-strategy}), the more commonly used beam search  \cite{https://doi.org/10.48550/arxiv.1606.02960} strategy leads to much less good results than sampling due to the lack of diversity induced by constant prediction (typical beams will look like $\sin(x), \sin(1.1x), \sin(0.9x), \ldots$). This provides us with a set of $\numbags \numbeam$ candidate solutions.

\paragraph{Inference time}

Our model inference speed has two sources: the forward passes described above on one hand (which can be parallelized up to memory limits of the GPU), and the refinements of candidate functions on the other (which are CPU-based and could also be parallelized, although we did not consider this option here). 

Since $\numbags \numbeam$ can become large, we rank candidate functions (according to their error on \emph{all} input points), get rid of redundant skeleton functions and keep the best $\numrefs$ candidates for the refinement step\footnote{Though these candidates are the best functions without refinement, there are no guarantees that these would be the best after refinement, especially as optimization is particularly prone to spurious local optimas.}. To speed up the refinement, we use a subset of at most $1024$ input points for the optimization. The parameters  $\numbags$, $\numbeam$ and $\numrefs$ can be used as cursors in the speed-accuracy tradeoff: in the experiments presented in Fig.~\ref{fig:pareto}, we selected $\numbags=100$, $\numbeam=10$, $\numrefs=10$.

\section{Results}

In this section, we present the results of our model. We begin by studying in-domain accuracy, then present results on out-of-domain datasets. 

\subsection{In-domain performance}
\label{sec:in-domain}
We report the in-domain performance of our models by evaluating them on a fixed validation set of 100,000 examples, generated as per Section~\ref{sec:generation}. Validation functions are uniformly spread out over three difficulty factors: number of unary operators, binary operators, and input dimension. For each function, we evaluate the performance of the model when presented $N=[50,100,150,200]$ input points $(x,y)$, and prediction accuracy is evaluated on $\ntest=200$ points sampled from a fresh instance of the multimodal distribution described in Section~\ref{sec:inputs}. 

We assess the performance of our model using two popular metrics: $R^2$-score~\cite{la2021contemporary} and accuracy to tolerance $\tau$~\cite{biggio2021neural,d2022deep}:
    \begin{align}
        R^{2}=1-\frac{\sum_{i}^{\ntest}\left(y_{i}-\hat{y}_{i}\right)^{2}}{\sum_{i}^{\ntest}\left(y_{i}-\bar{y}\right)^{2}},
        \quad \quad \quad
        \operatorname{Acc}_\tau = \mathds{1}\left(\max_{1\leq i \leq \ntest} \left\vert \frac{\hat y_i - y_i}{y_i}\right\vert \leq \tau \right),
        \label{eq:metrics}
    \end{align}
where $\mathds{1}$ is the indicator function.

$R^2$ is classically used in statistics, but it is unbounded, hence a single bad prediction can cause the average $R^2$ over a set of examples to be extremely bad. To circumvent this, we set $R^2=0$ upon pathological examples as in~\cite{la2021contemporary}(such examples occur in less that 1\% of cases)\footnote{Note that predicting the constant function $f = \bar y$ naturally yields an $R^2$ score of 0.}. 
The accuracy metric provides a better idea of the precision of the predicted expression as it depends on a desired tolerance threshold. However, due to the presence of the $\max$ operator, it is sensitive to outliers, and hence to the number of points considered at test time (more points entails a higher risk of outlier). To circumvent this, we discard the 5\% worst predictions, following~\cite{biggio2021neural}.

\paragraph{End-to-end outperforms skeleton}

\begin{table}{}
    \centering
    \begin{tabular}{c|c|c|c|c}
    \toprule
        Model                       & $R^2$ & $\text{Acc}_{0.1}$   & $\text{Acc}_{0.01}$ & $\text{Acc}_{0.001}$ \\
        \midrule
        Skeleton + BFGS                   & 0.43      & 0.40              & 0.27            & 0.17  \\
        E2E no BFGS                & 0.62      & 0.51              & 0.27            & 0.09  \\
        E2E + BFGS random init            & 0.44      & 0.44              & 0.30            & 0.19 \\
        E2E + BFGS model init                   & \textbf{0.68}      & \textbf{0.61}              & \textbf{0.44}  & \textbf{0.29}             \\
    \bottomrule
    \end{tabular}
    \vspace{.1cm}
    \caption{\textbf{Our approach outperforms the skeleton approach.} Metrics are computed over the $10,000$ examples of the evaluation set.}
    \label{tab:in-domain}
\end{table}
In Table~\ref{tab:in-domain}, we report the average in-domain results of our models. Without refinement, our E2E model outperforms the skeleton model trained under the same protocol in terms of low precision prediction ($R^2$ and $\text{Acc}_{0.1}$ metrics), but small errors in the prediction of the constants lead to lower performance at high precision ($\text{Acc}_{0.001}$ metric). The refinement procedure alleviates this issue significantly, inducing a three-fold increase in $\text{Acc}_{0.001}$ while also boosting other metrics. 

Initializing BFGS with the constants estimated in the E2E phase plays a crucial role: with random initialization, the BFGS step actually \emph{degrades} E2E performance. However, refinement with random initialization still achieves better results than the skeleton model: this suggests that the E2E model predicts skeletons better that the skeleton model.

\paragraph{Ablation}

\begin{figure*}
    \centering
    \includegraphics[width=\linewidth]{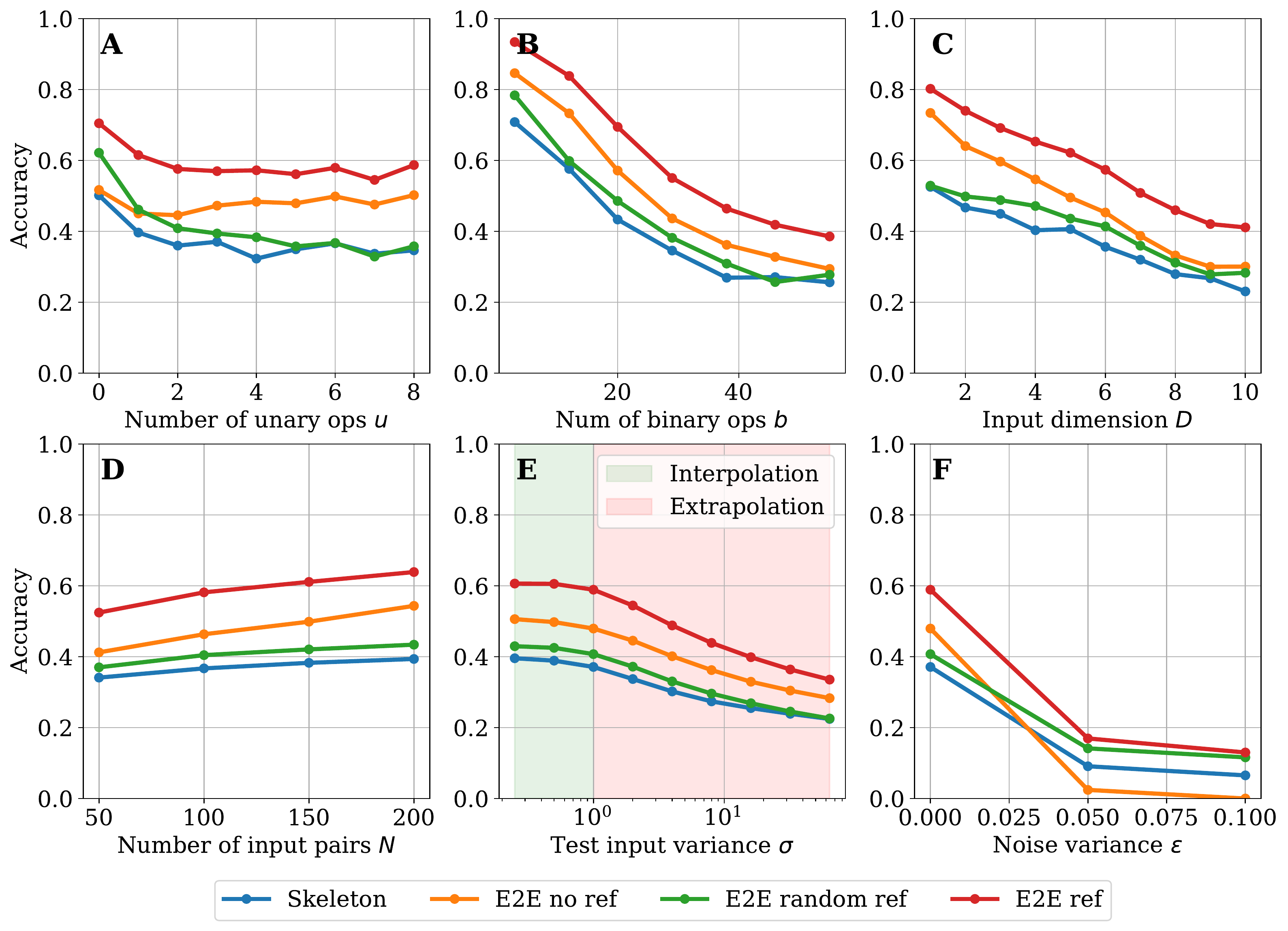}
    \caption{\textbf{Ablation over the function difficulty (top row) and input difficulty (bottom row).} We plot the accuracy at $\tau=0.1$ (Eq.~\ref{eq:metrics}), see App.~\ref{app:ablation} for the $R^2$ score. We distinguish four models: \textbf{\textcolor{C0}{skeleton}}, \textbf{\textcolor{C1}{E2E without refinement}}, \textbf{\textcolor{C2}{E2E with refinement from random guess}} and  \textbf{\textcolor{C3}{E2E with refinement}}.
    \textbf{A:} number of unary operators. \textbf{B:} number of binary operators. \textbf{C:} input dimension. \textbf{D:} Low-resource performance, evaluated by varying the number of input points. \textbf{E:} Extrapolation performance, evaluated by varying the variance of the inputs. \textbf{F:} Robustness to noise, evaluated by varying the multiplicative noise added to the labels.}
    \label{fig:in-domain}
\end{figure*}

Fig.~\ref{fig:in-domain}A,B,C presents an ablation over three indicators of formula difficulty (from left to right): number of unary operators, number of binary operators and input dimension. In all cases, increasing the factor of difficulty degrades performance, as one could expect. This may give the impression that our model does not scale well with the input dimension, but we show that our model scales in fact very well on out-of-domain datasets compared to concurrent methods (see Fig.~\ref{fig:ablation_input_dimension} of the Appendix).

Fig.~\ref{fig:in-domain}D shows how performance depends on the number of input points fed to the model, $N$. In all cases, performance increases, but much more signicantly for the E2E models than for the skeleton model, demonstrating the importance of having a lot of data to accurately predict the constants in the expression.

\paragraph{Extrapolation and robustness}

In Fig.~\ref{fig:in-domain}E, we examine the ability of our models to interpolate/extrapolate by varying the scale of the test points: instead of normalizing the test points to unit variance, we normalize them to a scale $\sigma$. As expected, performance degrades as we increase $\sigma$, however the extrapolation performance remains decent even very far away from the inputs ($\sigma=32$).

Finally, in Fig.~\ref{fig:in-domain}F, we examine the effect of corrupting the targets $y$ with a multiplicative noise of variance $\sigma$: $y \to y(1+\xi), \xi\sim \mathcal{N}(0,\varepsilon)$. The results reveal something interesting: without refinement, the E2E model is not robust to noise, and actually performs worse than the skeleton model at high noise. This shows how sensitive the Transformer is to the inputs when predicting constants. Refinement improves robustness significantly, but the initialization of constants to estimated values has less impact, since the prediction of constants is corrupted by the noise.

\subsection{Out-of-domain generalization}

We evaluate our method on the recently released benchmark SRBench\cite{la2021contemporary}. Its repository contains a set of $252$ regression datasets from the Penn Machine Learning Benchmark (PMLB)\cite{friedman2001greedy} in addition to $14$ open-source SR and ML baselines. The datasets consist in "ground-truth" problems where the true underlying function is known, as well as "black-box" problems which are more general regression datasets without an underlying ground truth.

We filter out problems from SRBench to only keep regression  problems with $\din \leq 10$ with continuous features; this results in $190$ regression datasets, splitted into $57$ black-box problems (combination of real-world and noisy, synthetic datasets), $119$ SR datasets from the Feynman~\cite{udrescu2020ai} and $14$ SR datasets from the ODE-Strogatz \cite{strogatz:2000} databases. Each dataset is split into $75\%$ training data and $25\%$ test data, on which performance is evaluated.

The overall performance of our models is illustrated in the Pareto plot of Fig.~\ref{fig:pareto}, where we see that on both types of problems, our model achieves performance close to state-of-the-art GP models such as Operon with a fraction of the inference time\footnote{Inference uses a single GPU for the forward pass of the Transformer.}. Impressively, our model outperforms all classic ML methods (e.g. XGBoost and Random Forests) on real-world problems with a lower inference time, and while outputting an interpretable formula. 

\begin{figure}[h]
    \centering
    \includegraphics[width=\linewidth]{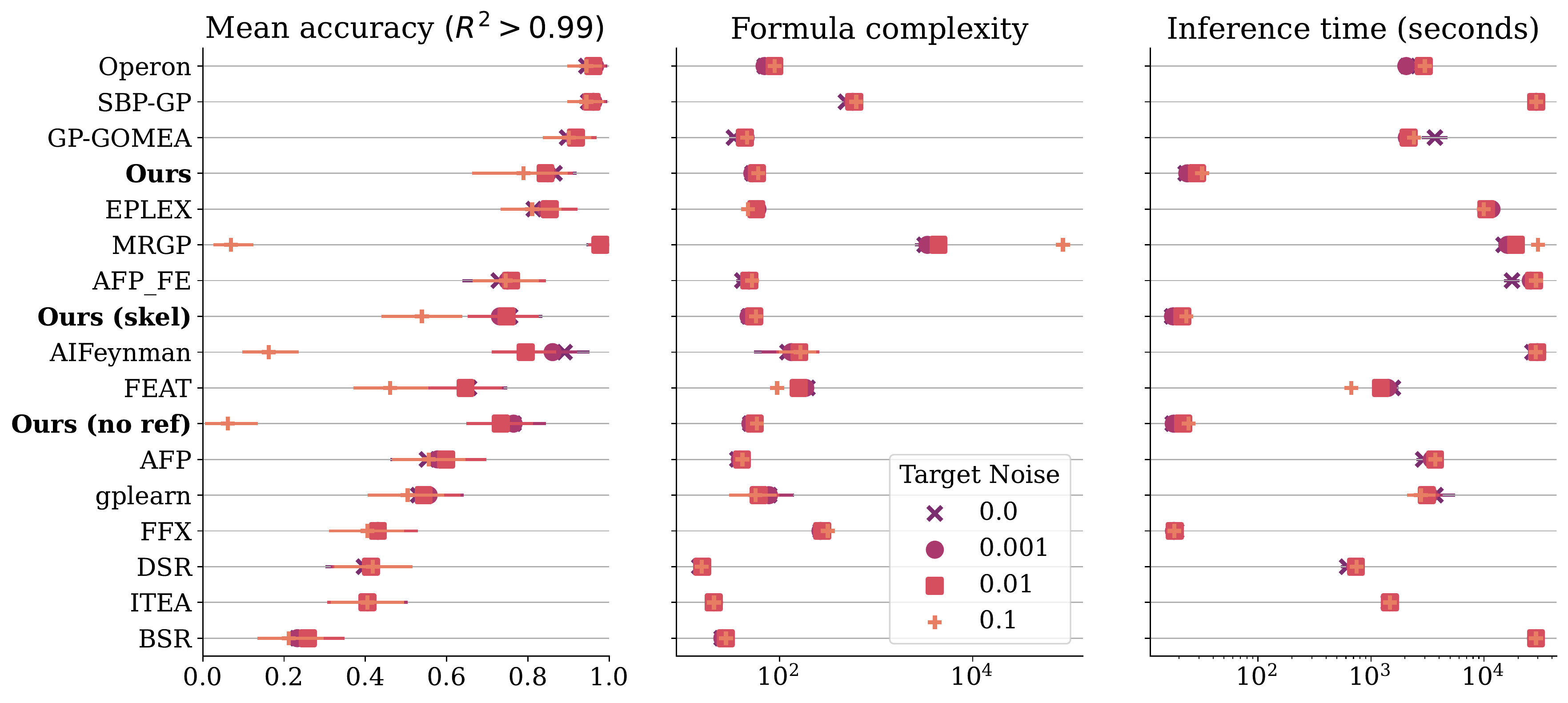}
    \caption{\textbf{Our model presents strong accuracy-speed-complexity tradeoffs, even in presence of noise.} Results are averaged over all 119 Feynman problems, for $10$ random seeds and three target noises each as shown in the legend. The accuracy is computed as the fraction of problems for which the $R^2$ score on test examples is above 0.99. Models are ranked according to the accuracy averaged over all target noise.}
    \label{fig:feynman_results}
\end{figure}

We provide more detailed results on Feynman problems in Fig. \ref{fig:feynman_results}, where we additionally plot the formula complexity, i.e. the number of nodes in the mathematical tree (see App.~\ref{app:results} for similar results on black-box and Strogatz problems).
Varying the noise applied to the targets noise, we see that our model displays similar robustness to state-of-the-art GP models. 

While the average accuracy or our model is only ranked fourth, it outputs formulas with lower complexity than the top 2 models (Operon and SBP-GP), which is an important criteria for SR problems: see App.~\ref{fig:pareto-size} for complexity-accuracy Pareto plots. To the best of our knowledge, our model is the first non-GP approach to achieve such competitive results for SR. 

\section*{Conclusion}

In this work, we introduced a competitive deep learning model for SR by using a novel numeric-symbolic approach. Through rigorous ablations, we showed that predicting the constants in an expression not only improves performance compared to predicting a skeleton, but can also serve as an informed initial condition for a solver to refine the value of the constants. 

Our model outperforms previous deep learning approaches by a margin on SR benchmarks, and scales to larger dimensions. Yet, the dimensions considered here remain moderate ($\din<10$): adapting to the truly high-dimensional setup is an interesting future direction, and will likely require qualitative changes in the data generation protocol. While our model narrows the gap between GP and DL based SR, closing the gap also remains a challenge for future work.

This work opens up a whole new range of applications for SR in fields which require real-time inference. We hope that the methods presented here may also serve as a toolbox for many future applications of Transformers for symbolic tasks.

%% file: appendix.tex
\section{Details on the training data}
\label{app:data}

In Tab.~\ref{tab:generator} we provide the detailed set of parameters used in our data generator.
The probabilities of the unary operators were selected to match the natural frequencies appearing in the Feynman dataset.

\begin{figure}[htb]
    \centering
    \includegraphics[width=\linewidth]{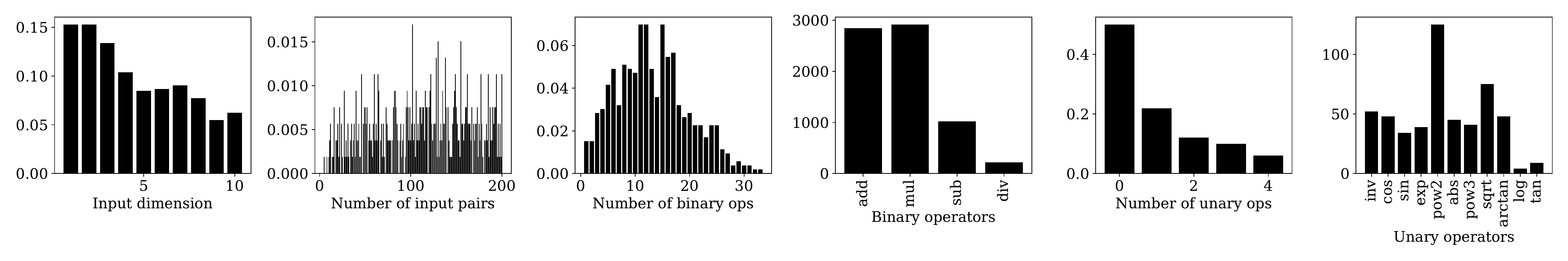}
    \caption{\textbf{Statistics of the synthetic data.} We calculated the latter on $10,000$ generated examples.}
    \label{fig:input_stats}
\end{figure}

In Fig.~\ref{fig:input_stats}, we show the statistics of the data generation.The number of expressions diminishes with the input dimension and number of unary operators because of the higher likelihood of generating out-of-domain inputs. One could easily make the distribution uniform by enforcing to retry as long as a valid example is not found, however we find empirically that having more easy examples than hard ones eases learning and provides better out-of-domain generalization, which is our ultimate goal.

\begin{figure}
    \centering
    \includegraphics[width=\linewidth]{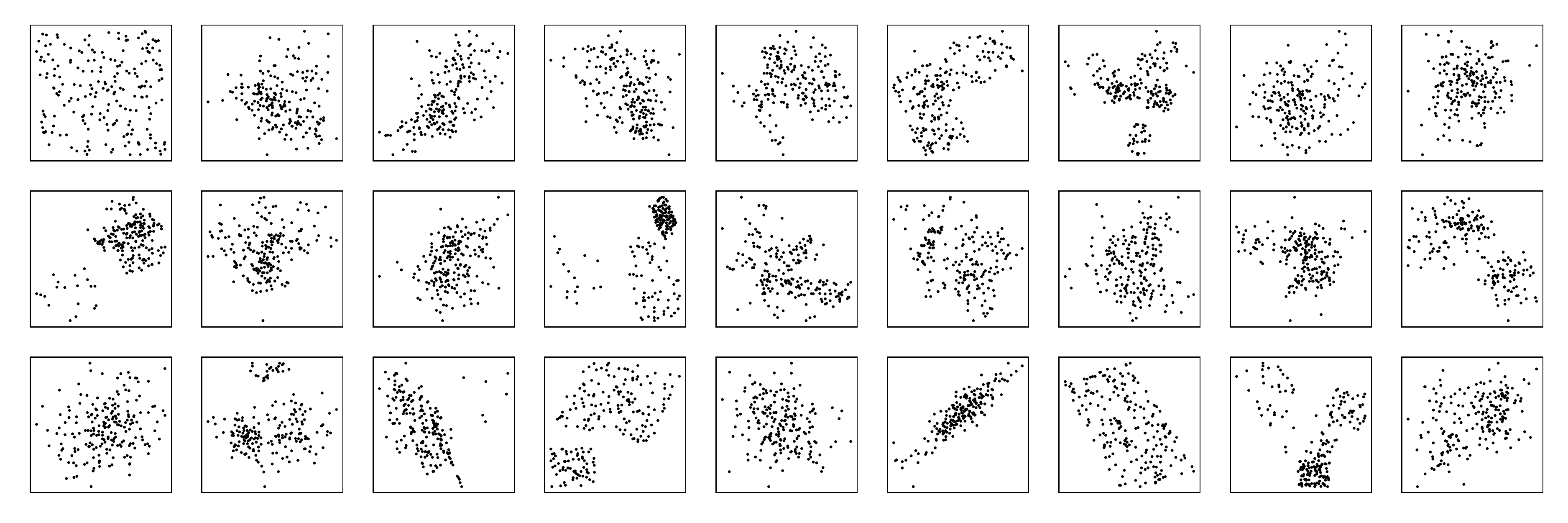}
\caption{\textbf{Diversity of the input distributions generated by the multimodal approach.} Here we show distributions obtained for $D=2$.}
    \label{fig:input_distrib}
\end{figure}

In Fig.~\ref{fig:input_distrib}, we show some examples of the input distributions generated by our multimodal approach. Notice the diversity of shapes obtained by this procedure.

\begin{table}[htb]
    \centering
    \begin{tabular}{c|c|c}
    \toprule
    Parameter & Description & Value\\
    \midrule
        $D_\text{max}$   & Max input dim & $10$ \\
        $\mathcal{D}_\text{aff}$ & Distrib of ($a$,$b$) & \makecell{sign $\sim \mathcal{U}\{-1,1\}$,\\ mantissa $\sim \mathcal{U}(0,1)$, \\ exponent $\sim \mathcal{U}(-2,2)$} \\
        \hline
        $b_\text{max}$   & Max binary ops & $5+D$ \\
        $O_b$ & Binary operators & add:1, sub:1, mul:1 \\
        \hline
        $u_\text{max}$   & Max unary ops & $5$\\
        $O_u$ & Unary operators & \makecell{inv:5, abs:1, sqr:3, sqrt:3, \\ sin:1, cos:1, tan:0.2, atan:0.2, \\ log:0.2, exp:1} \\
        \hline
        $N_\text{min}$ & Min number of points  & $10\din$ \\
        $N_\text{max}$ & Max number of points          & $200$ \\
        $k_{max}$ & Max num clusters & 10\\
    \bottomrule
    \end{tabular}
    \vspace{.3cm}
    \caption{\textbf{Parameters of our generator.}}
    \label{tab:generator}
\end{table}

\section{Attention maps}
\label{app:attention}

\begin{figure}[htb]
    \centering
    \begin{subfigure}[b]{.9\linewidth}
    \includegraphics[width=\columnwidth]{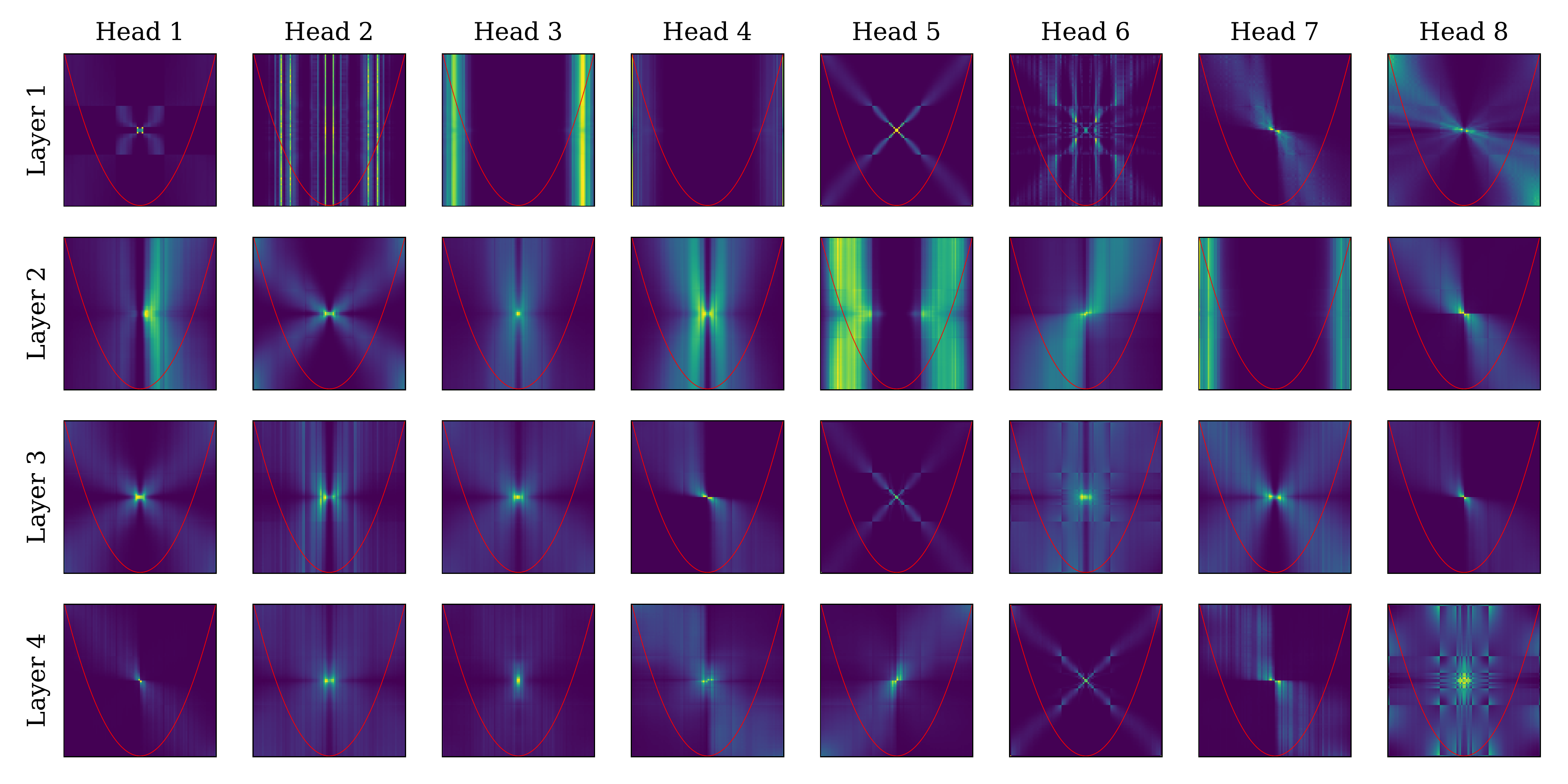}
    \caption{$f(x)=x^2$}
    \end{subfigure} 
    \begin{subfigure}[b]{.9\linewidth}
    \includegraphics[width=\columnwidth]{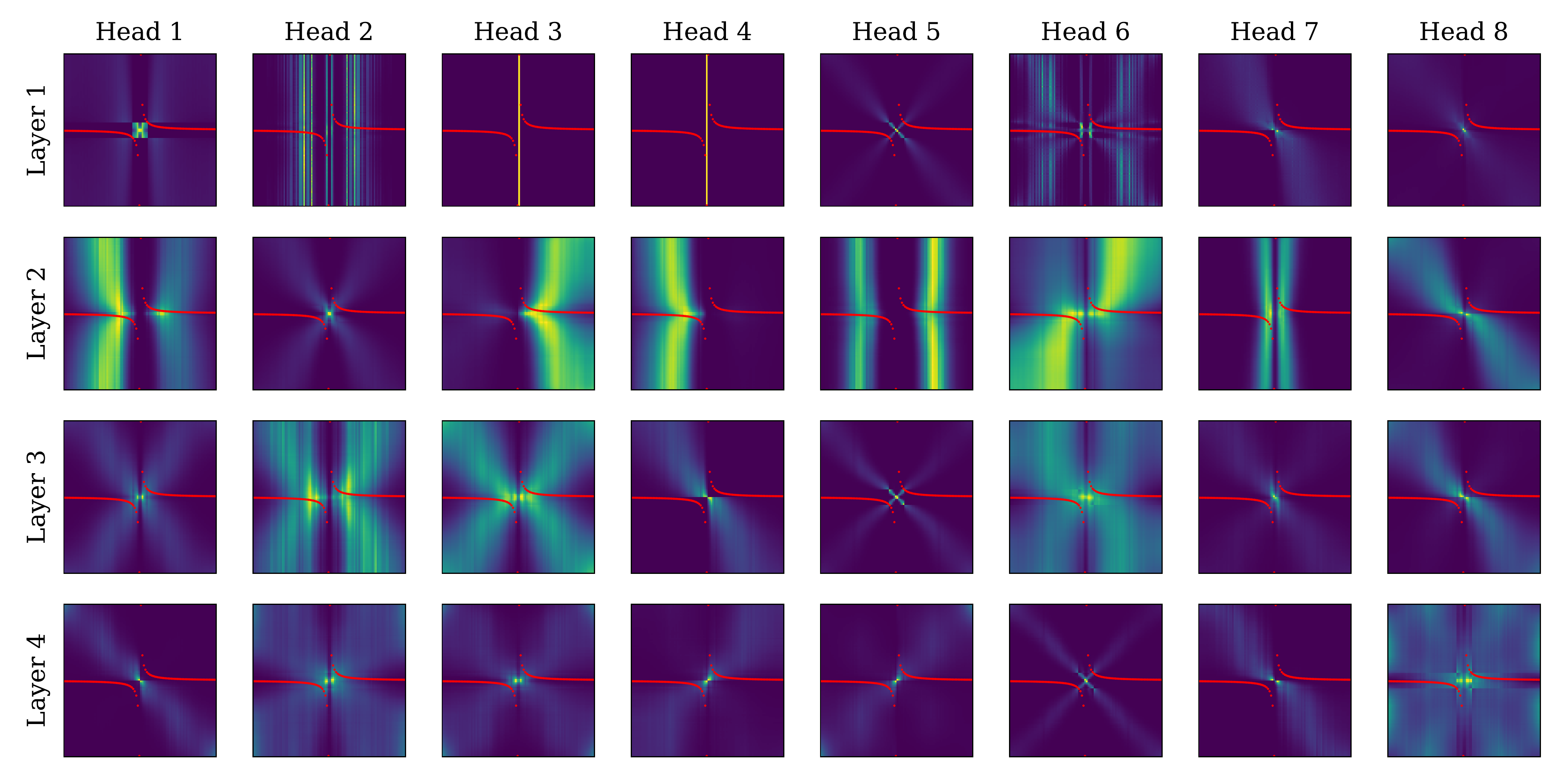}
    \caption{$f(x)=1/x$}
    \end{subfigure} 
    \begin{subfigure}[b]{.9\linewidth}
    \includegraphics[width=\columnwidth]{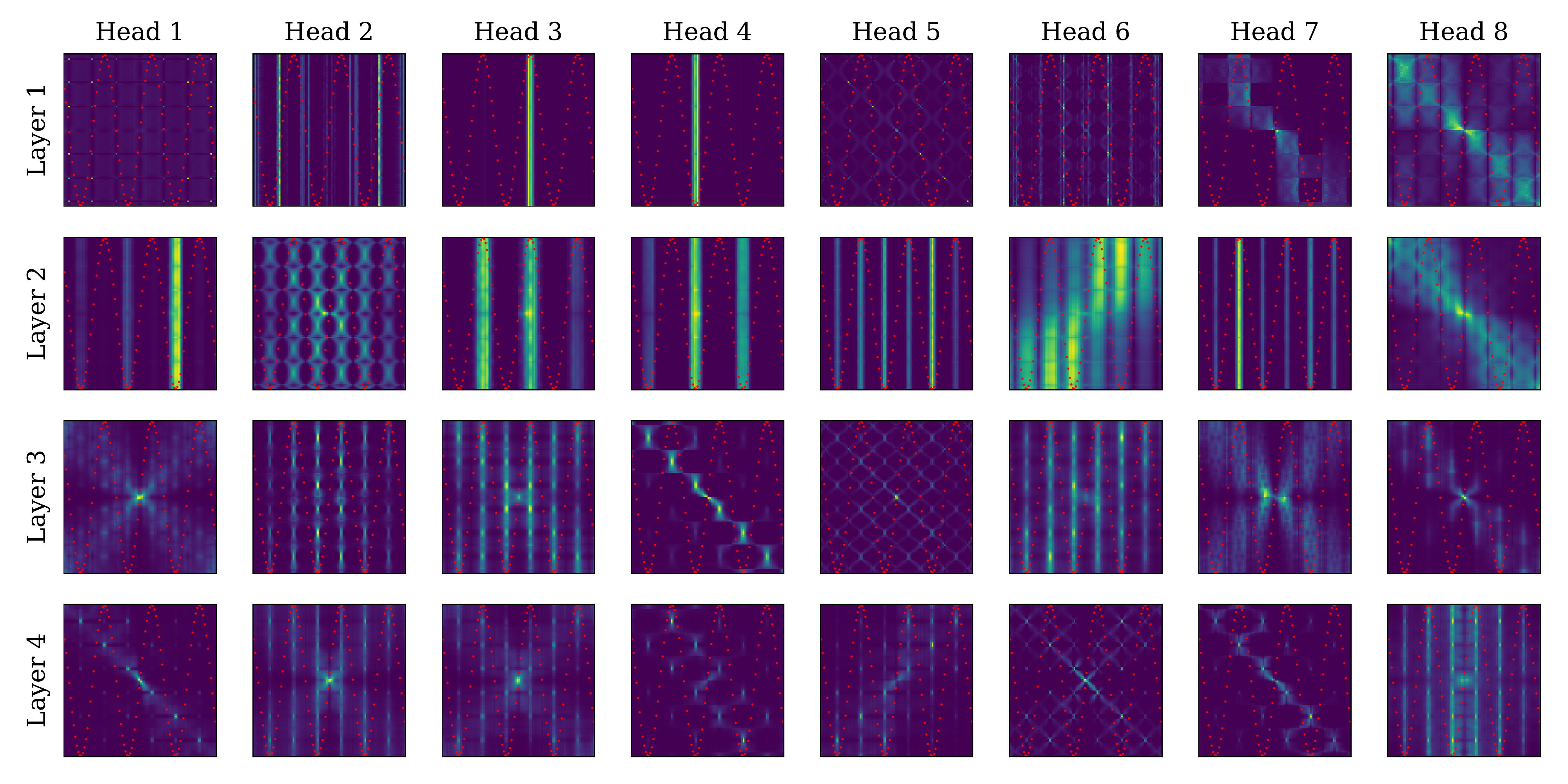}
    \caption{$f(x)=\sin (10x)$}
    \end{subfigure} 
    \caption{\textbf{Attention maps reveal distinctive features of the functions considered.} We presented the model 1-dimensional functions with 100 input points sorted in ascending order, in order to better visualize the attention. We plotted the self-attention maps of the first 8 (out of 16) heads of the Transformer encoder, across all four layers. We see very distinctive patterns appears: exploding areas for the exponential, the singularity at zero for the inverse function, and the periodicity of the sine function.}
    \label{fig:attention}
\end{figure}

A natural question is whether self-attention based architectures are optimally suited for symbolic regression tasks. In Fig.~\ref{fig:attention}, we show the attention maps produced by the encoder of our Transformer model, which contains 4 layers avec 16 attention heads (we only keep the first 8 for the sake of space). In order to make the maps readable, we consider one-dimensional inputs and sort them in ascending order.

The attention plots demonstrate the complementarity of the attention heads. Some focus on specific regions of the input, whereas others are more spread out. Some are concentrated along the diagonal (focusing on neighboring points), whereas others are concentrated along the anti-diagonal (focusing on far-away points. 

Most strikingly, the particular features of the functions studied clearly stand out in the attention plots. Focus, for example, on the 7th head of layer 2. For the exponential function, it focuses on the extreme points (near -1 and 1); for the inverse function, it focuses on the singularity around the origin; for the sine function, it reflects the periodicity, with evenly spaces vertical lines. The same phenomenology can be acrossed is several other heads.

\section{Does memorization occur?}
\label{app:memorization}

It is natural to ask the following question: due to the large amount of data seen during training, is our model simply memorizing the training set ? Answering this question involves computing the number of possible functions which can be generated. To estimate this number, calculating the number of possible skeleton $N_{s}$ is insufficient, since a given skeleton can give rise to very different functions according to the sampling of the constants, and even for a given choice of the constants, the input points $\{x\}$ can be sampled in many different ways.

\begin{figure}[htb]
    \centering
    \includegraphics[width=.5\columnwidth]{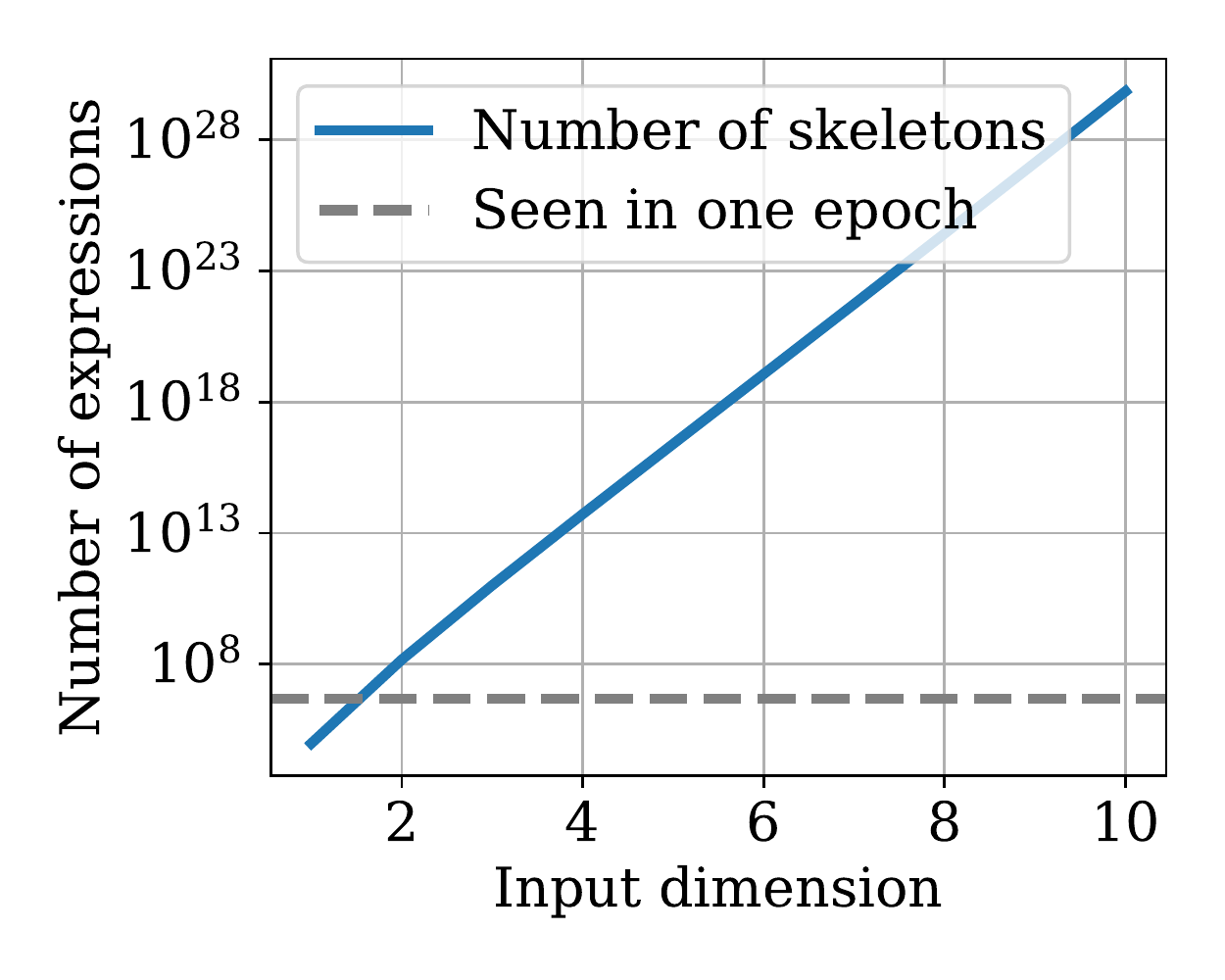}
    \caption{\textbf{Our models only see a small fraction of the possible expressions during training.} We report the number of possible skeletons for each number of operators. Even after a hundred epochs, our models have only seen a fraction of the possible expressions with more than 4 operators.}
    \label{fig:num_expressions}
\end{figure}

Nonetheless, we provide the lower bound $N_{s}$ as a function of the number of nodes in Fig.~\ref{fig:num_expressions}, using the equations provided in~\cite{lample2019deep}. For small expressions (up to four operators), the number of possible expressions is lower or similar to than the number of expressions encountered during training, hence one cannot exclude the possibility that some expressions were seen several times during training, but with different realizations due to the initial conditions. However, for larger expressions, the number of possibilities is much larger, and one can safely assume that the expressions encountered at test time have not been seen during training.

\section{Additional in-domain results}
\label{app:ablation}
\begin{figure*}
    \centering
    \includegraphics[width=\linewidth]{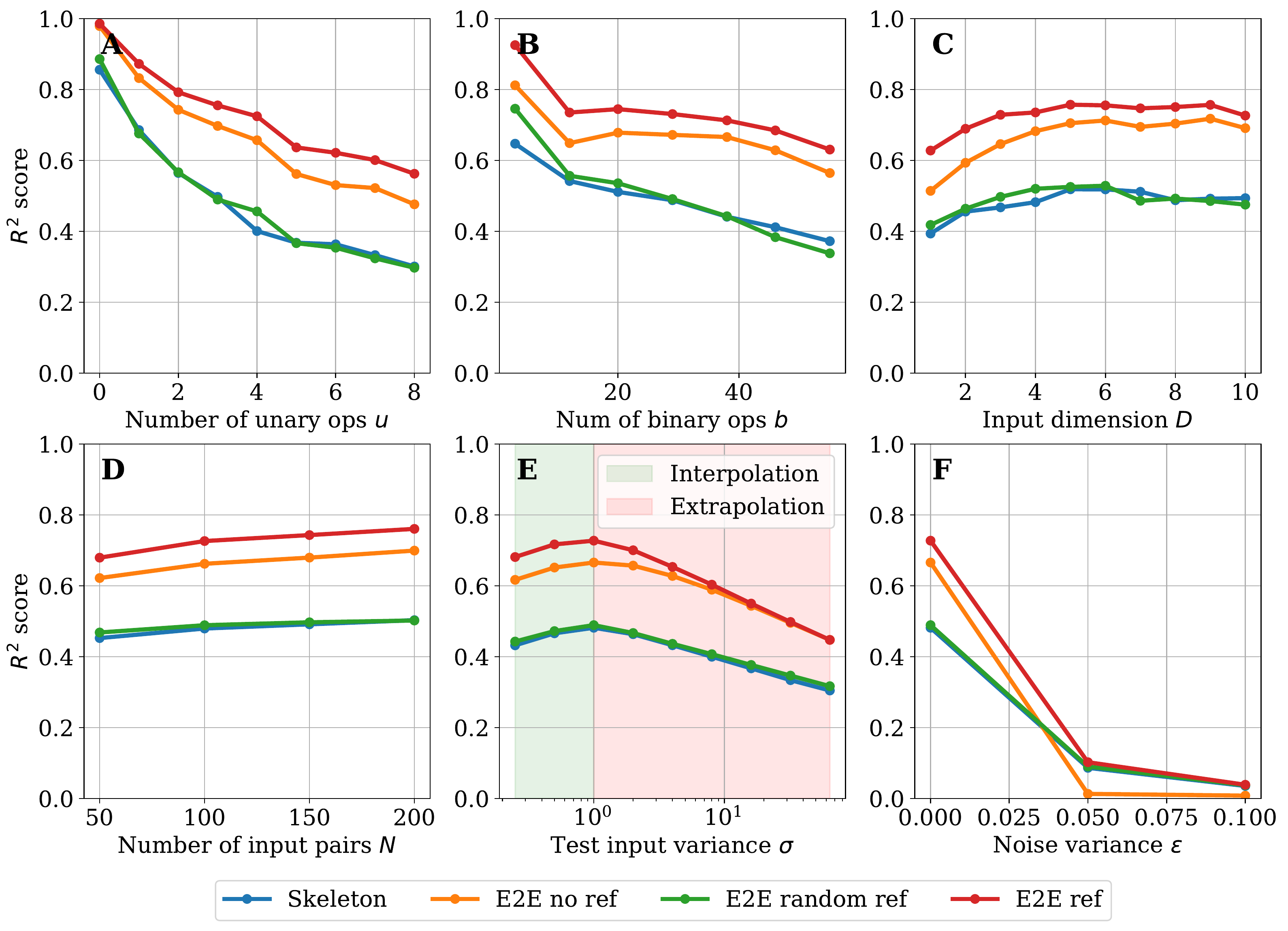}
    \caption{\textbf{Ablation over the function difficulty (top row) and input difficulty (bottom row).} We plot the $R^2$ score (Eq.~\ref{eq:metrics}). \textbf{A:} number of unary operators. \textbf{B:} number of binary operators. \textbf{C:} input dimension. \textbf{D:} Low-resource performance, evaluated by varying the number of input points. \textbf{E:} Extrapolation performance, evaluated by varying the variance of the inputs. \textbf{F:} Robustness to noise, evaluated by varying the multiplicative noise added to the labels.}
    \label{fig:in-domain-r2}
\end{figure*}

Fig.~\ref{fig:in-domain-r2}, we present a similar ablation as Fig.~\ref{fig:in-domain} of the main text but using the $R^2$ score as metric rather than accuracy. 

\section{Additional out-of-domain results}
\label{app:results}

\paragraph{Complexity-accuracy }

\begin{figure}
    \includegraphics[width=\linewidth]{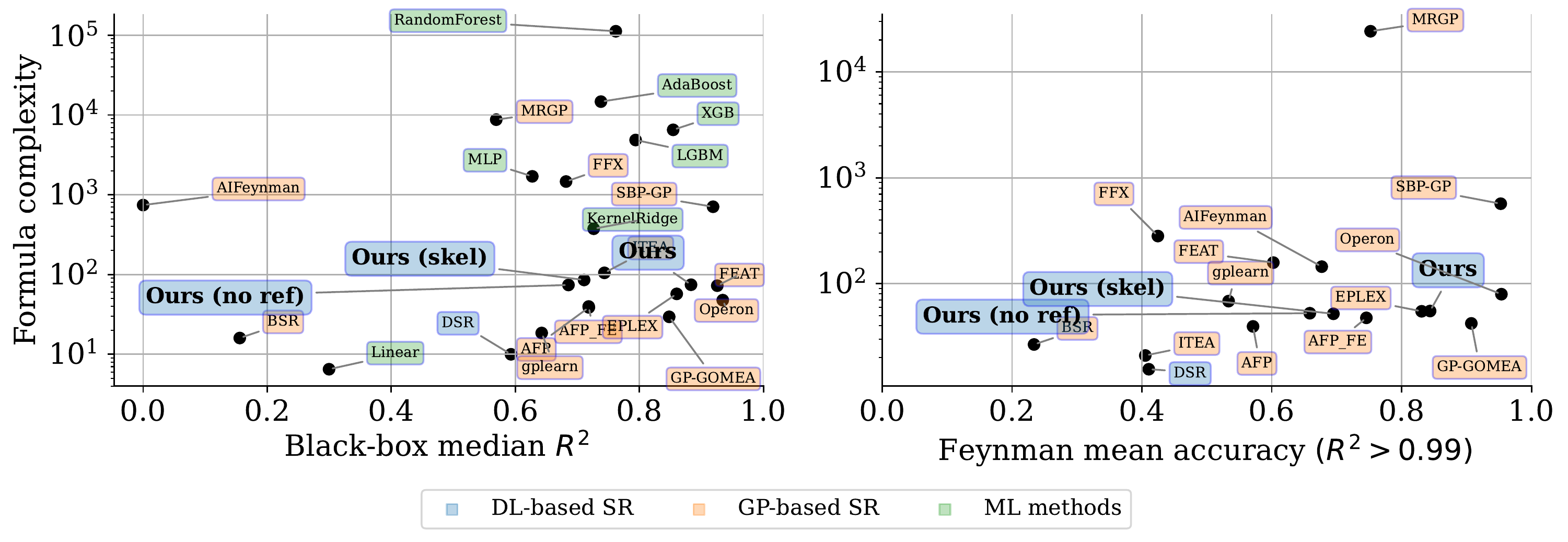}
  \captionof{figure}{\textbf{Complexity-accuracy pareto plot.} Pareto plot comparing the average test performance and formula complexity of our models with baselines provided by the SRbench benchmark~\cite{la2021contemporary}, both on Feynman SR problems~\cite{udrescu2020ai} and black-box regression problems. We use colors to distinguish three families of models: deep-learning based SR, genetic programming-based SR and classic machine learning methods (which do not provide an interpretable solution).}
  \label{fig:pareto-size}
  \end{figure}

In Fig.~\ref{fig:pareto-size}, we display a Pareto plot comparing accuracy and formula complexity on SRBench datasets.

\paragraph{Jin benchmark}

\begin{figure}[htb]
    \centering
    \begin{subfigure}[b]{.49\linewidth}
    \includegraphics[width=\linewidth]{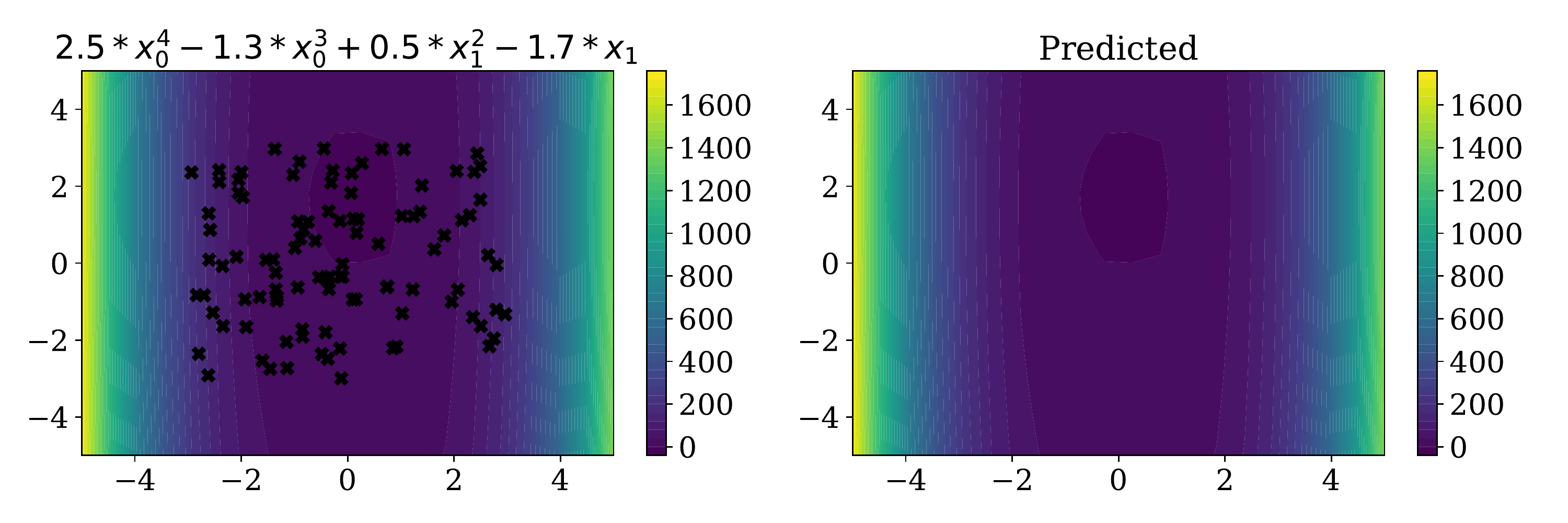}
    \caption{Jin-1}
    \end{subfigure} 
    \begin{subfigure}[b]{.49\linewidth}
    \includegraphics[width=\linewidth]{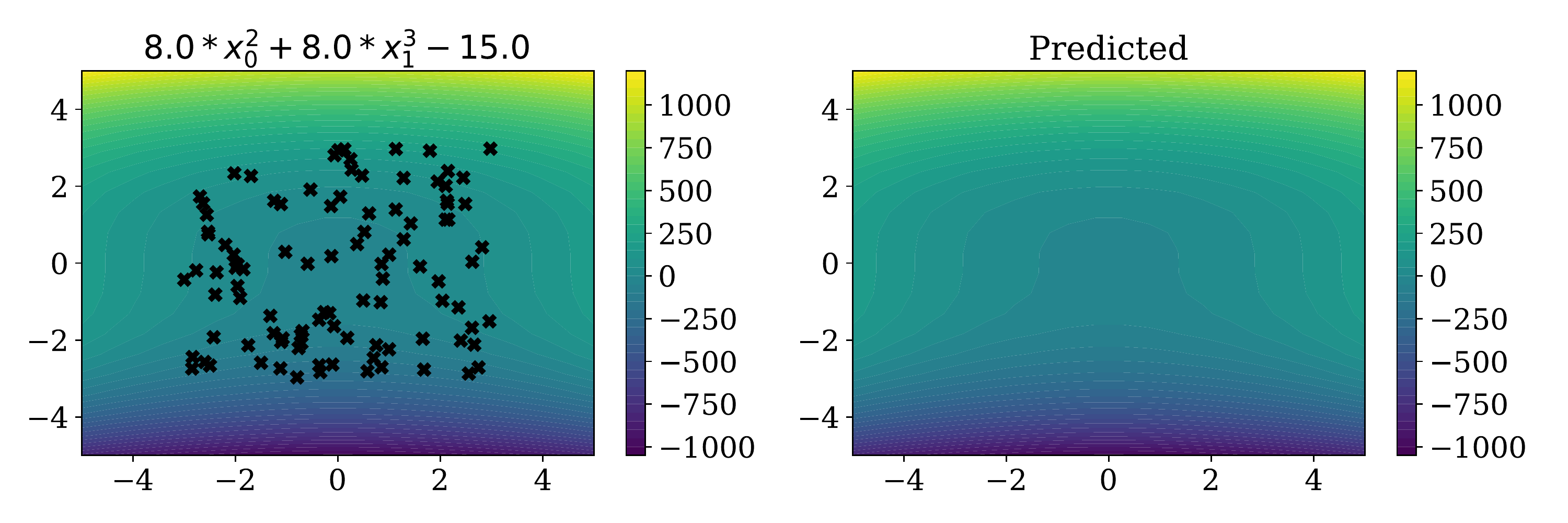}
    \caption{Jin-2}
    \end{subfigure} 
    \begin{subfigure}[b]{.49\linewidth}
    \includegraphics[width=\linewidth]{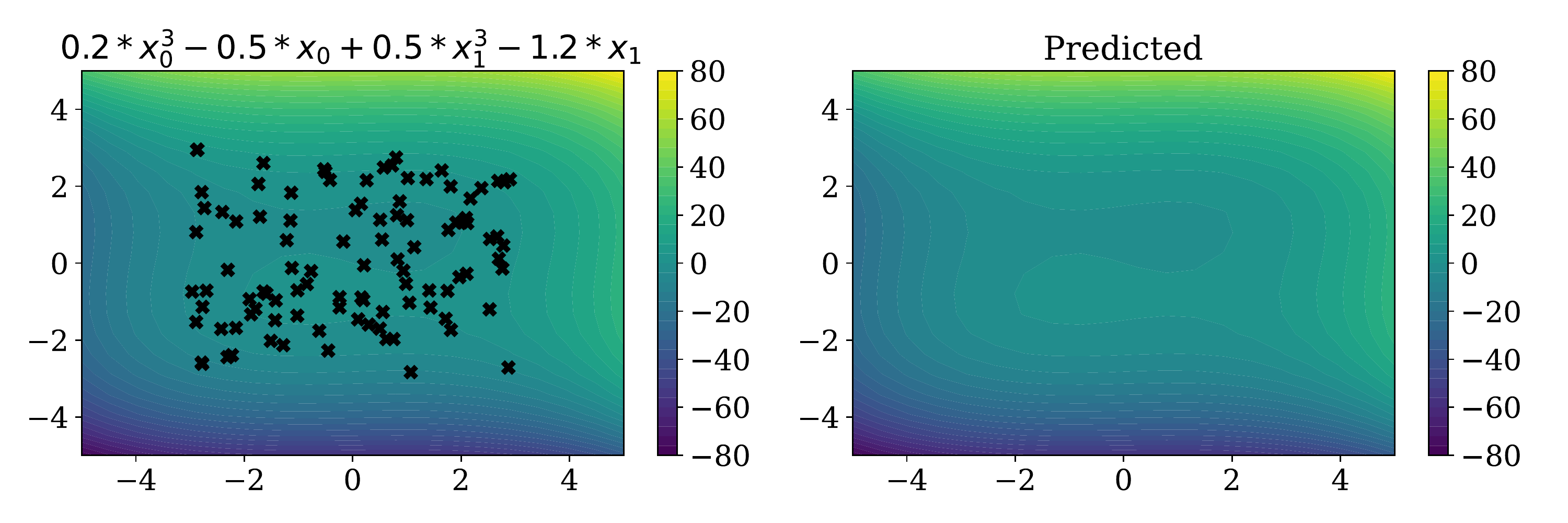}
    \caption{Jin-3}
    \end{subfigure} 
    \begin{subfigure}[b]{.49\linewidth}
    \includegraphics[width=\linewidth]{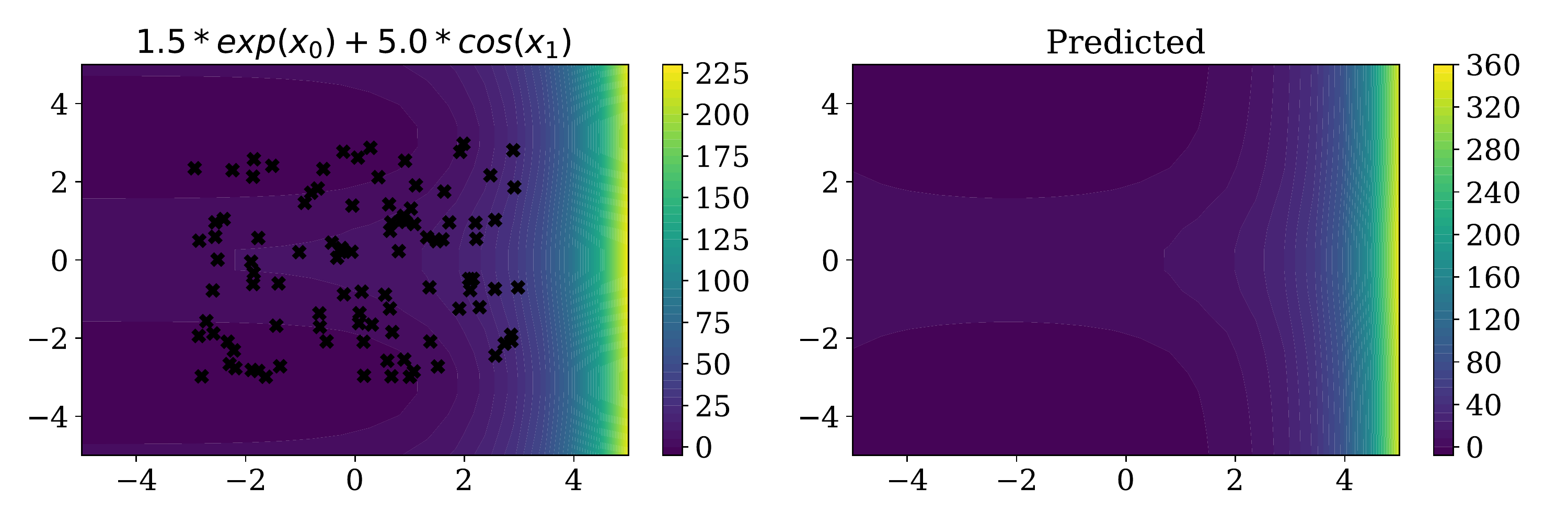}
    \caption{Jin-4}
    \end{subfigure} 
    \begin{subfigure}[b]{.49\linewidth}
    \includegraphics[width=\linewidth]{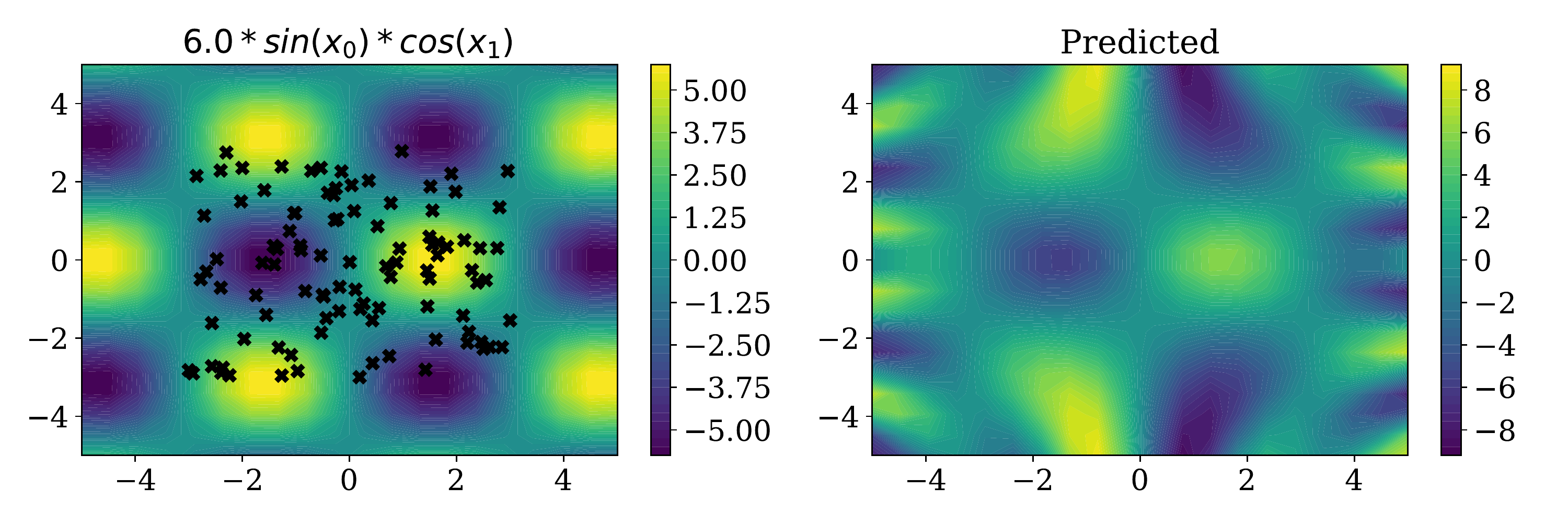}
    \caption{Jin-5}
    \end{subfigure} 
    \begin{subfigure}[b]{.49\linewidth}
    \includegraphics[width=\linewidth]{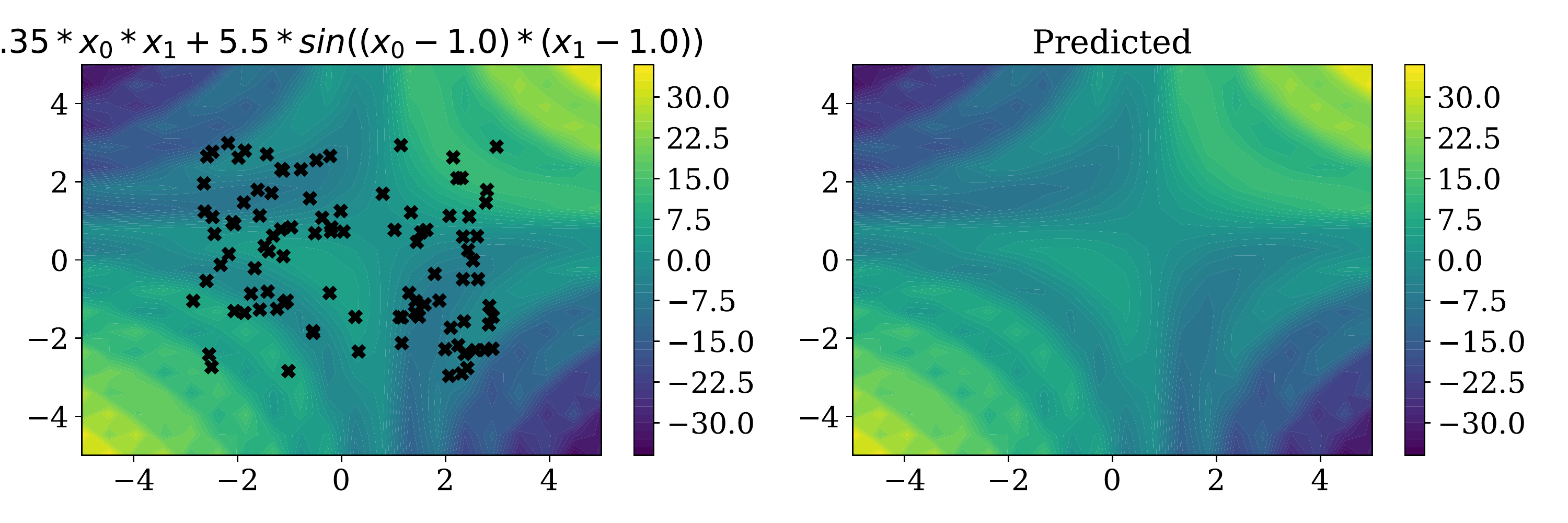}
    \caption{Jin-6}
    \end{subfigure} 

    \caption{\textbf{Illustration of our model on a few benchmark datasets from the litterature.} We show the prediction of our model on six 2-dimensional datasets presented in~\cite{jin2020bayesian} and used as a comparison point in a few recent works~\cite{mundhenk2021symbolic}. The input points are marked as black crosses. Our model retrieves the correct expression in all but one of the cases: in Jin5, the prediction matches the input points correctly, but extrapolates badly.}
    \label{fig:jin}
\end{figure}

In Fig.~\ref{fig:jin}, we show the predictions of our model on the functions provided in~\cite{jin2020bayesian}. Our model gets all of them correct except for one.

\paragraph{Black-box datasets}

\begin{figure}[h]
    \centering
    \includegraphics[width=\linewidth]{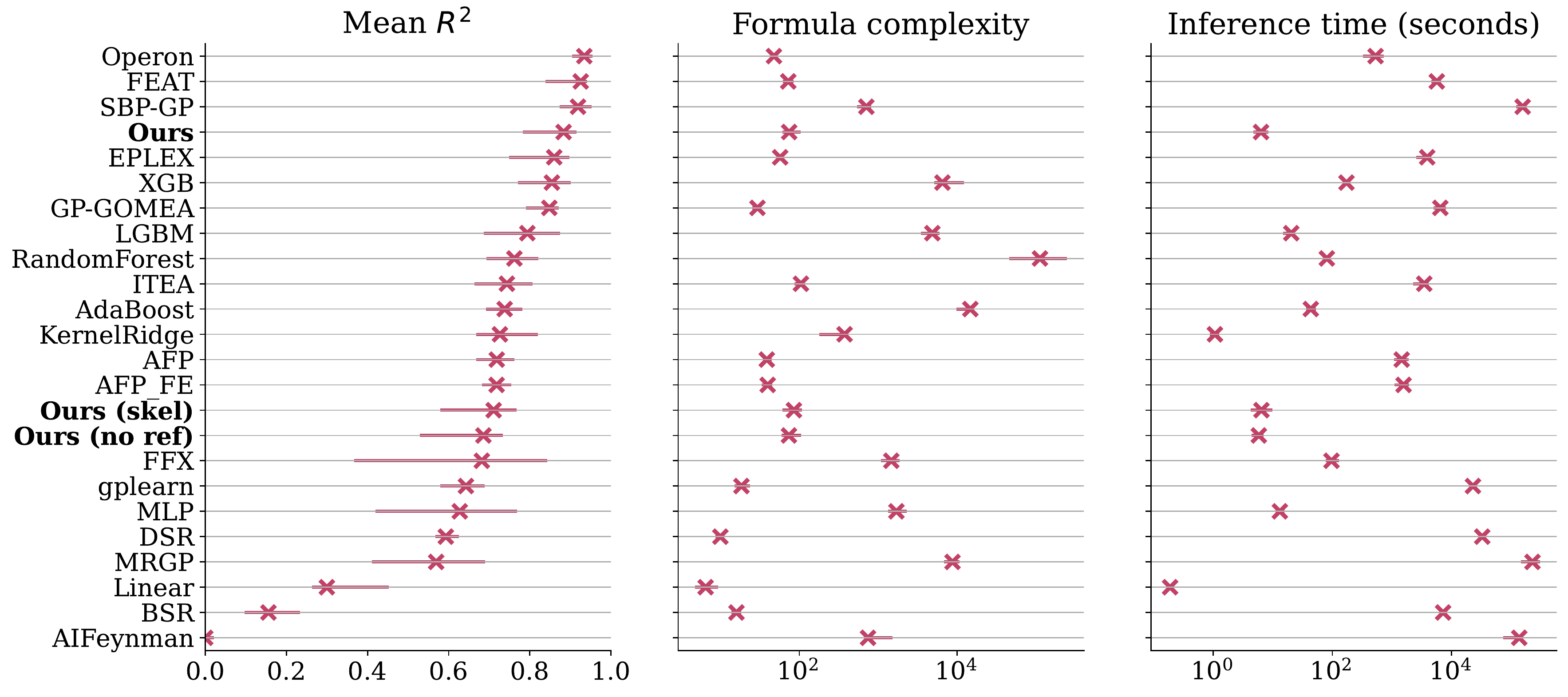}
    \caption{\textbf{Performance metrics on black-box datasets.} }
    \label{fig:black_box_results}
\end{figure}

In Fig.~\ref{fig:black_box_results}, we display the results of our model on the black-box problems of SRBench.

\paragraph{Strogatz datasets}

\begin{figure}[htb]
     \centering
     \includegraphics[width=\linewidth]{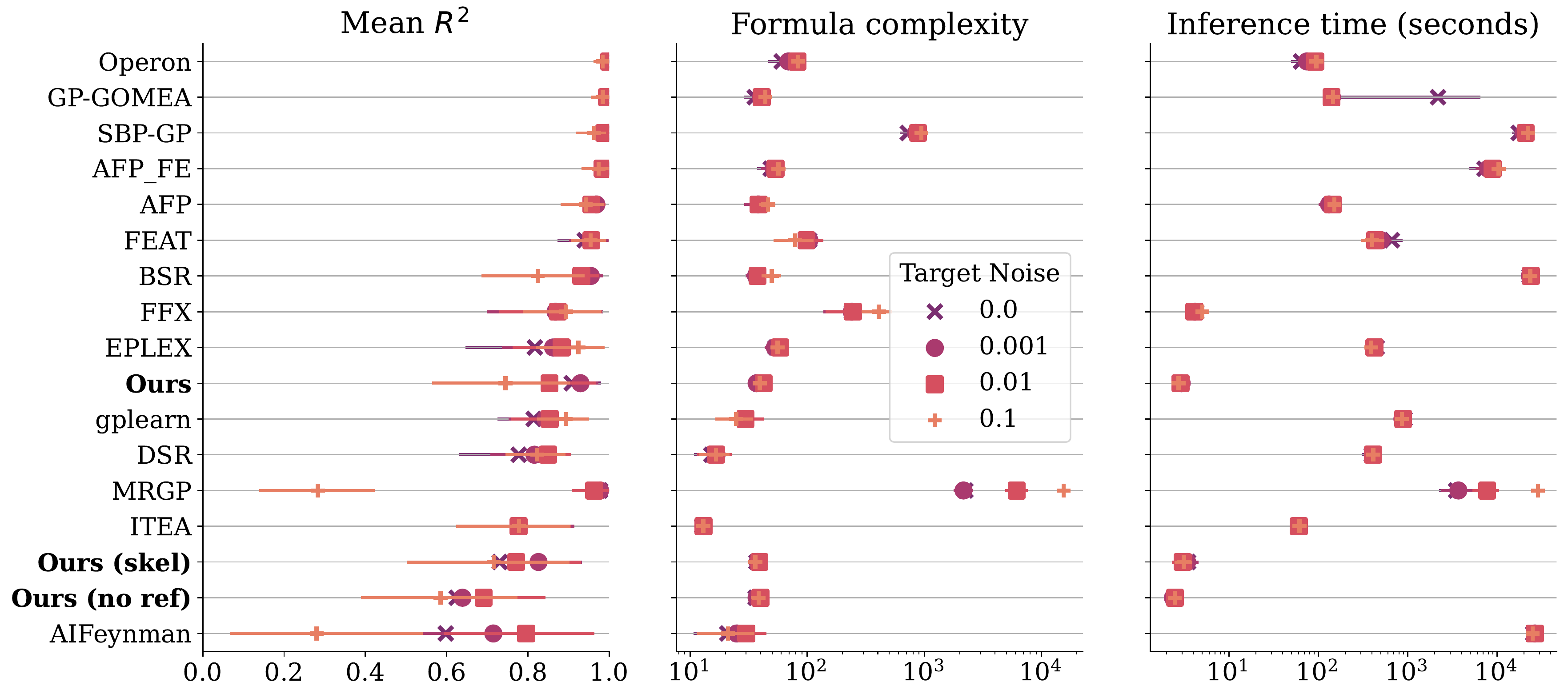}
     \caption{\textbf{Performance metrics on Strogatz datasets.}}
     \label{fig:strogatz_results}
 \end{figure}

Each of the 14 datasets from the ODE-Strogatz benchmark is the trajectory of a
2-state system following a first-order ordinary differential equation (ODE). Therefore, the input data has a very particular, time-ordered distribution, which differs significantly from that seen at train time. Unsurprisingly, Fig.~\ref{fig:strogatz_results} shows that our model performs somewhat less well to this kind of data in comparison with GP-based methods. 

\begin{figure}[h]
    \centering
    \begin{subfigure}[b]{\linewidth}
    \includegraphics[width=\linewidth]{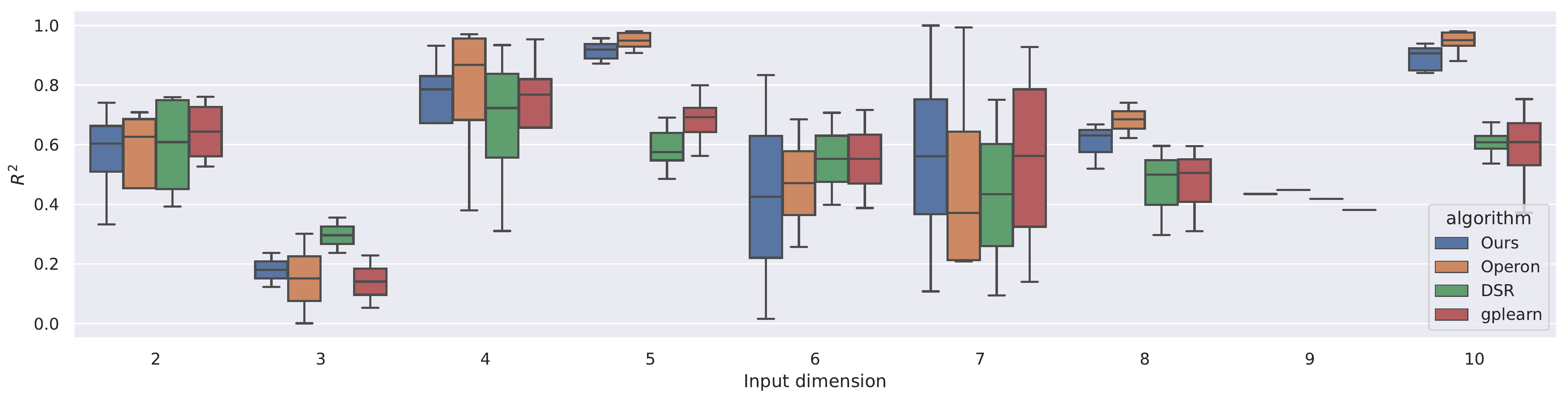}
    \caption{Black-box}
    \end{subfigure}
    \begin{subfigure}[b]{\linewidth}
    \includegraphics[width=\linewidth]{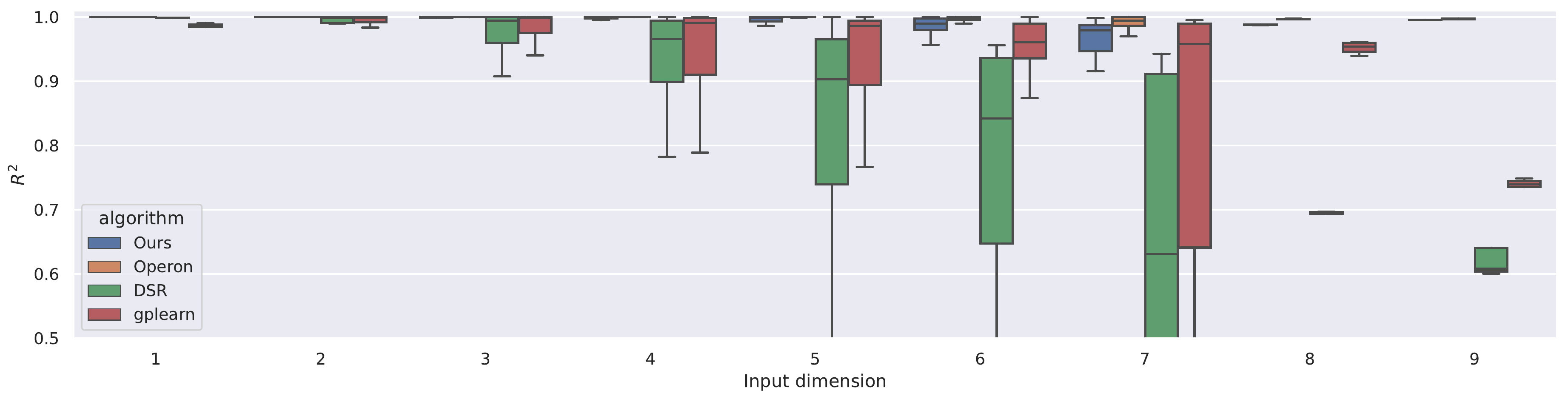}
    \caption{Feynman}
    \end{subfigure}
    \caption{\textbf{Performance metrics on SRBench, separated by input dimension.} }
    \label{fig:ablation_input_dimension}
\end{figure}

\paragraph{Ablation on input dimension} In Fig.~\ref{fig:ablation_input_dimension}, we show how the performance of our model depends on the dimensionality of the inputs on Feynamn and black-box datasets. 

\paragraph{Ablation on decoding strategy}
\label{app:beam}

\begin{figure}[h]
    \centering
    \includegraphics[width=.5\linewidth]{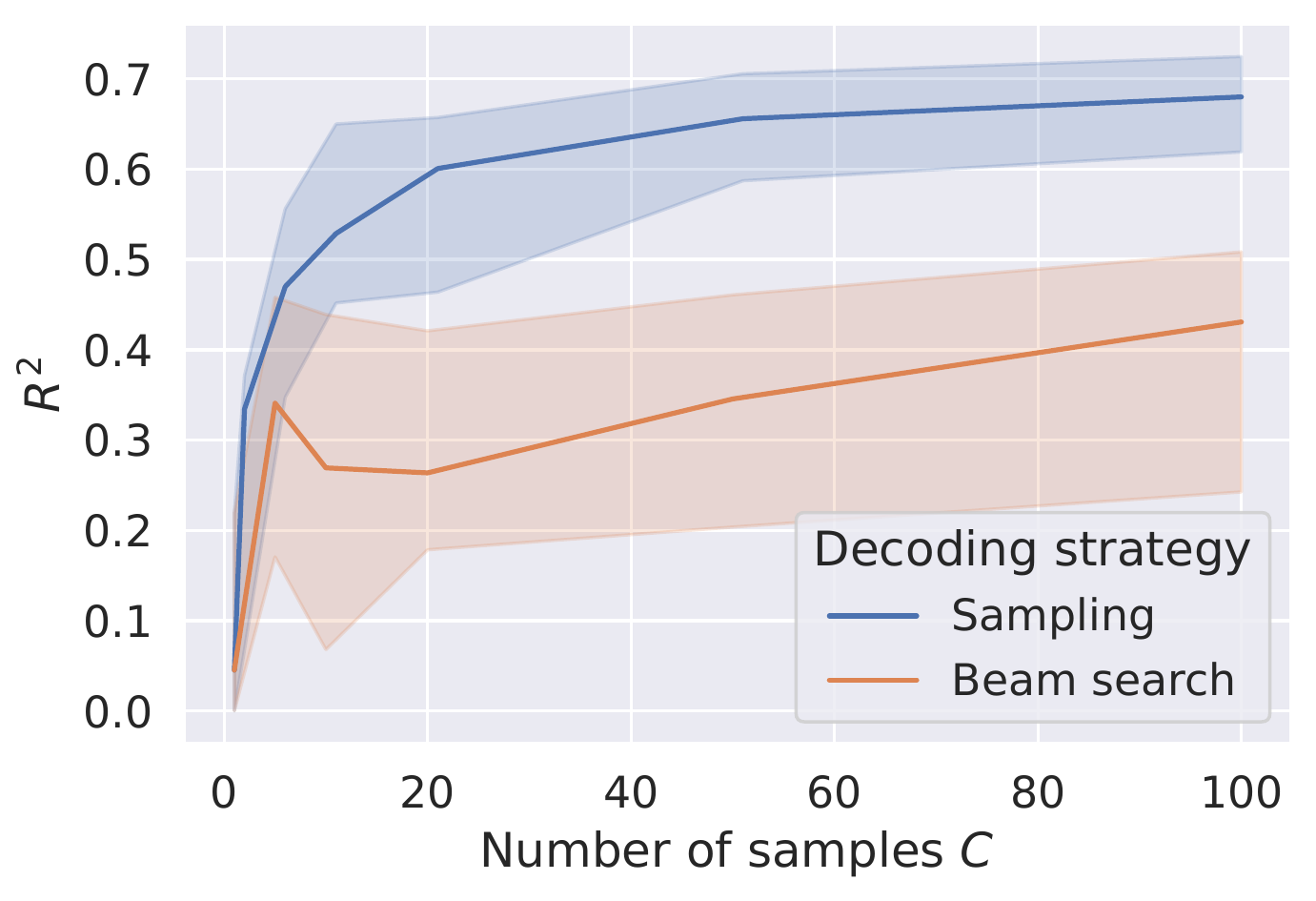}
    \caption{\textbf{Median $R^2$ of our method without refinement on black-box datasets when $B=1$, varying the number of decoded function samples.} The beam search \cite{https://doi.org/10.48550/arxiv.1606.02960} used in \cite{biggio2021neural} leads to low-diversity candidates in our setup due to expressions differing only by small modifications of the coefficients.}
    \label{fig:decoding-strategy}
\end{figure}

In Fig.~\ref{fig:decoding-strategy}, we display the difference in performance using two decoding strategies.

%% file: refs.bib
@article{lample2019deep,
  title={Deep learning for symbolic mathematics},
  author={Lample, Guillaume and Charton, Fran{\c{c}}ois},
  journal={arXiv preprint arXiv:1912.01412},
  year={2019}
}

@article{petersen2019deep,
  title={Deep symbolic regression: Recovering mathematical expressions from data via risk-seeking policy gradients},
  author={Petersen, Brenden K and Larma, Mikel Landajuela and Mundhenk, T Nathan and Santiago, Claudio P and Kim, Soo K and Kim, Joanne T},
  journal={arXiv preprint arXiv:1912.04871},
  year={2019}
}

@article{valipour2021symbolicgpt,
  title={SymbolicGPT: A Generative Transformer Model for Symbolic Regression},
  author={Valipour, Mojtaba and You, Bowen and Panju, Maysum and Ghodsi, Ali},
  journal={arXiv preprint arXiv:2106.14131},
  year={2021}
}

@inproceedings{vaswani2017attention,
  title={Attention is all you need},
  author={Vaswani, Ashish and Shazeer, Noam and Parmar, Niki and Uszkoreit, Jakob and Jones, Llion and Gomez, Aidan N and Kaiser, {\L}ukasz and Polosukhin, Illia},
  booktitle={Advances in neural information processing systems},
  pages={5998--6008},
  year={2017}
}

@inproceedings{sahoo2018learning,
  title={Learning equations for extrapolation and control},
  author={Sahoo, Subham and Lampert, Christoph and Martius, Georg},
  booktitle={International Conference on Machine Learning},
  pages={4442--4450},
  year={2018},
  organization={PMLR}
}

@article{charton2020learning,
  title={Learning advanced mathematical computations from examples},
  author={Charton, Fran{\c{c}}ois and Hayat, Amaury and Lample, Guillaume},
  journal={arXiv preprint arXiv:2006.06462},
  year={2020}
}

@misc{jin2020bayesian,
      title={Bayesian Symbolic Regression}, 
      author={Ying Jin and Weilin Fu and Jian Kang and Jiadong Guo and Jian Guo},
      year={2020},
      eprint={1910.08892},
      archivePrefix={arXiv},
      primaryClass={stat.ME}
}

@book{strogatz:2000,
  author = {Strogatz, Steven H.},
  publisher = {Westview Press},
  title = {Nonlinear Dynamics and Chaos: With Applications to Physics, Biology, Chemistry and Engineering},
  year = 2000
}

@misc{biggio2021neural,
      title={Neural Symbolic Regression that Scales}, 
      author={Luca Biggio and Tommaso Bendinelli and Alexander Neitz and Aurelien Lucchi and Giambattista Parascandolo},
      year={2021},
      eprint={2106.06427},
      archivePrefix={arXiv},
      primaryClass={cs.LG}
}

@misc{udrescu2020ai,
      title={AI Feynman: a Physics-Inspired Method for Symbolic Regression}, 
      author={Silviu-Marian Udrescu and Max Tegmark},
      year={2020},
      eprint={1905.11481},
      archivePrefix={arXiv},
      primaryClass={physics.comp-ph}
}

@misc{mundhenk2021symbolic,
      title={Symbolic Regression via Neural-Guided Genetic Programming Population Seeding}, 
      author={T. Nathan Mundhenk and Mikel Landajuela and Ruben Glatt and Claudio P. Santiago and Daniel M. Faissol and Brenden K. Petersen},
      year={2021},
      eprint={2111.00053},
      archivePrefix={arXiv},
      primaryClass={cs.NE}
}

@article{friedman2001greedy,
  title={Greedy function approximation: a gradient boosting machine},
  author={Friedman, Jerome H},
  journal={Annals of statistics},
  pages={1189--1232},
  year={2001},
  publisher={JSTOR}
}

@article{charton2021linear,
  title={Linear algebra with transformers},
  author={Charton, Fran{\c{c}}ois},
  journal={arXiv preprint arXiv:2112.01898},
  year={2021}
}

@Misc{functorch2021,
  author =       {Horace He, Richard Zou},
  title =        {functorch: JAX-like composable function transforms for PyTorch},
  howpublished = {\url{https://github.com/pytorch/functorch}},
  year =         {2021}
}

@misc{sympytorch,
  author = {Patrick Kidger},
  title = {SympyTorch},
  year = {2021},
  publisher = {GitHub},
  journal = {GitHub repository},
  howpublished = {\url{https://github.com/patrick-kidger/sympytorch}}
}

@misc{https://doi.org/10.48550/arxiv.1606.02960,
  doi = {10.48550/ARXIV.1606.02960},
  
  url = {https://arxiv.org/abs/1606.02960},
  
  author = {Wiseman, Sam and Rush, Alexander M.},
  
  title = {Sequence-to-Sequence Learning as Beam-Search Optimization},
  
  publisher = {arXiv},
  
  year = {2016},
  
  copyright = {arXiv.org perpetual, non-exclusive license}
}

@article{d2022deep,
  title={Deep Symbolic Regression for Recurrent Sequences},
  author={d'Ascoli, St{\'e}phane and Kamienny, Pierre-Alexandre and Lample, Guillaume and Charton, Fran{\c{c}}ois},
  journal={arXiv preprint arXiv:2201.04600},
  year={2022}
}

@article{la2021contemporary,
  title={Contemporary symbolic regression methods and their relative performance},
  author={La Cava, William and Orzechowski, Patryk and Burlacu, Bogdan and de Franca, Fabricio Olivetti and Virgolin, Marco and Jin, Ying and Kommenda, Michael and Moore, Jason H},
  journal={arXiv preprint arXiv:2107.14351},
  year={2021}
}

@inproceedings{landajuela2021discovering,
  title={Discovering symbolic policies with deep reinforcement learning},
  author={Landajuela, Mikel and Petersen, Brenden K and Kim, Sookyung and Santiago, Claudio P and Glatt, Ruben and Mundhenk, Nathan and Pettit, Jacob F and Faissol, Daniel},
  booktitle={International Conference on Machine Learning},
  pages={5979--5989},
  year={2021},
  organization={PMLR}
}

@article{garnelo2016towards,
  title={Towards deep symbolic reinforcement learning},
  author={Garnelo, Marta and Arulkumaran, Kai and Shanahan, Murray},
  journal={arXiv preprint arXiv:1609.05518},
  year={2016}
}

@incollection{schmidt2011age,
  title={Age-fitness pareto optimization},
  author={Schmidt, Michael and Lipson, Hod},
  booktitle={Genetic programming theory and practice VIII},
  pages={129--146},
  year={2011},
  publisher={Springer}
}

@article{schmidt2009distilling,
  title={Distilling free-form natural laws from experimental data},
  author={Schmidt, Michael and Lipson, Hod},
  journal={science},
  volume={324},
  number={5923},
  pages={81--85},
  year={2009},
  publisher={American Association for the Advancement of Science}
}

@article{la2018learning,
  title={Learning concise representations for regression by evolving networks of trees},
  author={La Cava, William and Singh, Tilak Raj and Taggart, James and Suri, Srinivas and Moore, Jason H},
  journal={arXiv preprint arXiv:1807.00981},
  year={2018}
}

@incollection{mcconaghy2011ffx,
  title={FFX: Fast, scalable, deterministic symbolic regression technology},
  author={McConaghy, Trent},
  booktitle={Genetic Programming Theory and Practice IX},
  pages={235--260},
  year={2011},
  publisher={Springer}
}

@article{virgolin2021improving,
  title={Improving model-based genetic programming for symbolic regression of small expressions},
  author={Virgolin, Marco and Alderliesten, Tanja and Witteveen, Cees and Bosman, Peter AN},
  journal={Evolutionary computation},
  volume={29},
  number={2},
  pages={211--237},
  year={2021},
  publisher={MIT Press One Rogers Street, Cambridge, MA 02142-1209, USA journals-info~…}
}

@article{de2021interaction,
  title={Interaction--Transformation Evolutionary Algorithm for Symbolic Regression},
  author={de Fran{\c{c}}a, Fabricio Olivetti and Aldeia, Guilherme Seidyo Imai},
  journal={Evolutionary computation},
  volume={29},
  number={3},
  pages={367--390},
  year={2021},
  publisher={MIT Press One Rogers Street, Cambridge, MA 02142-1209, USA journals-info~…}
}

@article{martius2016extrapolation,
  title={Extrapolation and learning equations},
  author={Martius, Georg and Lampert, Christoph H},
  journal={arXiv preprint arXiv:1610.02995},
  year={2016}
}

@inproceedings{arnaldo2014multiple,
  title={Multiple regression genetic programming},
  author={Arnaldo, Ignacio and Krawiec, Krzysztof and O'Reilly, Una-May},
  booktitle={Proceedings of the 2014 Annual Conference on Genetic and Evolutionary Computation},
  pages={879--886},
  year={2014}
}

@article{kommenda2020genetic,
	Author = {Kommenda, Michael and Burlacu, Bogdan and Kronberger, Gabriel and Affenzeller, Michael},
	Da = {2020/09/01},
	Doi = {10.1007/s10710-019-09371-3},
	Id = {Kommenda2020},
	Isbn = {1573-7632},
	Journal = {Genetic Programming and Evolvable Machines},
	Number = {3},
	Pages = {471--501},
	Title = {Parameter identification for symbolic regression using nonlinear least squares},
	Ty = {JOUR},
	Url = {https://doi.org/10.1007/s10710-019-09371-3},
	Volume = {21},
	Year = {2020}}

@inproceedings{virgolin2019linear,
author = {Virgolin, Marco and Alderliesten, Tanja and Bosman, Peter A. N.},
title = {Linear Scaling with and within Semantic Backpropagation-Based Genetic Programming for Symbolic Regression},
year = {2019},
isbn = {9781450361118},
publisher = {Association for Computing Machinery},
address = {New York, NY, USA},
url = {https://doi.org/10.1145/3321707.3321758},
doi = {10.1145/3321707.3321758},
booktitle = {Proceedings of the Genetic and Evolutionary Computation Conference},
pages = {1084–1092},
numpages = {9},
location = {Prague, Czech Republic},
series = {GECCO '19}
}

@article{Cranmer2020DiscoveringSM,
  title={Discovering Symbolic Models from Deep Learning with Inductive Biases},
  author={M. Cranmer and Alvaro Sanchez-Gonzalez and Peter W. Battaglia and Rui Xu and Kyle Cranmer and David N. Spergel and Shirley Ho},
  journal={ArXiv},
  year={2020},
  volume={abs/2006.11287}
}

@article{Archiga2021AcceleratingUO,
  title={Accelerating Understanding of Scientific Experiments with End to End Symbolic Regression},
  author={Nikos Ar{\'e}chiga and Francine Chen and Yan-Ying Chen and Yanxia Zhang and Rumen Iliev and Heishiro Toyoda and Kent Lyons},
  journal={ArXiv},
  year={2021},
  volume={abs/2112.04023}
}

@inproceedings{Butter2021BackTT,
  title={Back to the Formula -- LHC Edition},
  author={Anja Butter and Tilman Plehn and Nathalie Soybelman and Johann Brehmer},
  year={2021}
}

@article{Udrescu2021SymbolicPD,
  title={Symbolic Pregression: Discovering Physical Laws from Raw Distorted Video},
  author={Silviu-Marian Udrescu and Max Tegmark},
  journal={Physical review. E},
  year={2021},
  volume={103 4-1},
  pages={
          043307
        }
}

@article{guimera2020bayesian,
  title={A Bayesian machine scientist to aid in the solution of challenging scientific problems},
  author={Guimer{\`a}, Roger and Reichardt, Ignasi and Aguilar-Mogas, Antoni and Massucci, Francesco A and Miranda, Manuel and Pallar{\`e}s, Jordi and Sales-Pardo, Marta},
  journal={Science advances},
  volume={6},
  number={5},
  pages={eaav6971},
  year={2020},
  publisher={American Association for the Advancement of Science}
}

@article{polu2020generative,
  title={Generative language modeling for automated theorem proving},
  author={Polu, Stanislas and Sutskever, Ilya},
  journal={arXiv preprint arXiv:2009.03393},
  year={2020}
}

@article{hahn2020teaching,
  title={Teaching temporal logics to neural networks},
  author={Hahn, Christopher and Schmitt, Frederik and Kreber, Jens U and Rabe, Markus N and Finkbeiner, Bernd},
  journal={arXiv preprint arXiv:2003.04218},
  year={2020}
}
